\documentclass[preprint,12pt]{elsarticle}

\usepackage{amsmath,amsfonts}
\usepackage{algorithmic}
\usepackage{algorithm}
\usepackage{array}
\usepackage[caption=false,font=normalsize,labelfont=sf,textfont=sf]{subfig}
\usepackage{textcomp}
\usepackage{stfloats}
\usepackage{url}
\usepackage{verbatim}
\usepackage{graphicx}
\usepackage{hyperref}
\usepackage{booktabs}
\usepackage{multirow}
\usepackage{makecell}
\usepackage{pifont}
\usepackage[table]{xcolor}
\usepackage{amssymb}
\usepackage{amsmath}
\usepackage{makecell}
\usepackage{tabularx}
\usepackage{amssymb}

\journal{Neural Networks}

\begin{document}

\begin{frontmatter}


\title{MHAF-YOLO: Multi-Branch Heterogeneous Auxiliary Fusion YOLO for accurate object detection}

% \author[1]{Huihuang~Zhang}
% \author[1]{Haigen~Hu\corref{cor1}} 
% \author[1]{Qi~Chen}
% \author[1]{Qianwei~Zhou}
% \author[1]{Qiu~Guan}
\author[1]{Zhiqiang Yang}
\author[1]{Qiu Guan\textsuperscript{*} }
\author[1]{Zhongwen Yu}
\author[1]{Xinli Xu}
\author[1]{Haixia Long}
\author[2]{Sheng Lian}
\author[1]{Haigen Hu}
\author[1]{ Ying Tang\textsuperscript{*} }

\address[1]{ Zhejiang University of Technology, Hangzhou,310014, P.R. China}
\address[2]{Fuzhou University, Fuzhou, 350108, P.R. China}
% \affiliation[label1]{organization={},

% \affiliation[label2]{organization={},

\cortext[cor1]{Corresponding author, Email: gq@zjut.edu.cn}

%% Abstract
\begin{abstract}
% Due to its effective capabilities of multi-scale feature fusion, Path Aggregation FPN (PAFPN) is widely adopted in YOLO detectors. However, it struggles to sufficiently integrate high-level semantic information with low-level spatial information, making it challenging to handle real-world applications that demand high precision for multi-scale object detection. In response, this paper proposes a novel model named MHAF-YOLO, which features a versatile neck known as Multi-Branch Auxiliary FPN (MAFPN) within a new object detection framework. The MAFPN includes two key modules: the Superficial Assisted Fusion (SAF) module and the Advanced Assisted Fusion (AAF) module. SAF module bridges the backbone and the neck by fusing shallow features, effectively transferring crucial low-level spatial information from the backbone to the neck with high fidelity. Meanwhile, AAF module integrates multi-scale feature information at deeper levels of the neck, transmitting a richer variety of gradient information to the output layer and further strengthening the learning capacity of model.

Due to the effective multi-scale feature fusion capabilities of the Path Aggregation FPN (PAFPN), it has become a widely adopted component in YOLO-based detectors. However, PAFPN struggles to integrate high-level semantic cues with low-level spatial details, limiting its performance in real-world applications, especially with significant scale variations.
% However, PAFPN struggles to fully integrate high-level semantic cues with low-level spatial details, limiting its performance in real-world applications that demand precise multi-scale object detection, particularly in scenarios with drastic scale variations.
% However, PAFPN struggles to fully integrate high-level semantic cues with low-level spatial details, limiting its performance in real-world applications that demand precise multi-scale object detection. 
In this paper, we propose MHAF-YOLO, a novel detection framework featuring a versatile neck design called the Multi-Branch Auxiliary FPN (MAFPN), which consists of two key modules: the Superficial Assisted Fusion (SAF) and Advanced Assisted Fusion (AAF). The SAF bridges the backbone and the neck by fusing shallow features, effectively transferring crucial low-level spatial information with high fidelity. Meanwhile, the AAF integrates multi-scale feature information at deeper neck layers, delivering richer gradient information to the output layer and further enhancing the model’s learning capacity.
To complement MAFPN, we introduce the Global Heterogeneous Flexible Kernel Selection (GHFKS) mechanism and the Re-parameterized Heterogeneous Multi-Scale (RepHMS) module to enhance feature fusion. 
RepHMS is globally integrated into the network, utilizing GHFKS to select larger convolutional kernels for various feature layers, expanding the vertical receptive field and capturing contextual information across spatial hierarchies. Locally, it optimizes convolution by processing both large and small kernels within the same layer, broadening the lateral receptive field and preserving crucial details for detecting smaller targets.
% Globally, RepHMS is strategically integrated throughout the network, leveraging GHFKS to select larger convolutional kernels for various feature layers. This strategy expands the model’s vertical receptive field, effectively capturing contextual information across diverse spatial hierarchies. Locally, RepHMS optimizes convolution operations by simultaneously processing large and small kernels within the same feature layer. This approach broadens the lateral receptive field, enabling the network to capture multi-scale information while preserving essential details for detecting smaller targets.
% Using the small version of MHAF-YOLO as an example, our model achieves 48.9\% AP on the COCO dataset with only 7.1M learnable parameters. This represents a 24.4\% reduction in parameter count compared to YOLO11s, alongside a 1.9\% improvement in performance. Aside from this, our model has demonstrated outstanding performance in both instance segmentation and rotated object detection.
The small version of MHAF-YOLO achieves 48.9\% AP on COCO with just 7.1M parameters, a 24.4\% reduction compared to YOLO11s, while improving performance by 1.9\%. Additionally, our model has demonstrated outstanding performance and generalization in both instance segmentation and rotated object detection.
% To complement the MAFPN, the Global Heterogeneous Kernel Selection (GHKS) mechanism and the Re-parameterized Heterogeneous Multi-Scale (RepHMS) module are strategically implemented to refine the feature fusion process further. 
% From a global perspective, RepHMS is deployed throughout the network, utilizing the GHKS strategy to adaptively select larger convolutional kernels in different feature layers, thereby enhancing the model's vertical receptive field. Concurrently, the RepHMS module optimizes convolution operations at a local level. It facilitates the parallel operation of large and small convolutional kernels within the same feature layer, broadening the lateral receptive field. This innovative approach not only extends the network's capability to perceive wider areas but also meticulously preserves the details of smaller targets.
The source code of this work is available at: \href{https://github.com/yang-0201/MHAF-YOLO}{https://github.com/yang-0201/MHAF-YOLO}. 
\end{abstract}


% \begin{highlights}
% \end{highlights}

%% Keywords

\end{frontmatter}



\section{Introduction}
In recent years, a variety of algorithms have been developed to implement real-time object detection with high performance. Among them, a series of YOLO algorithms~\cite{yolo1,yolo9000,yolov3,yolov4,yolov5,yolox,yolov6,yolov7,yolov8,damo,rtmdet,ppyoloe,goldyolo,yoloms,yolov9, mafyolo, yolov10, yolo11, yolov12}, from YOLOv1 to YOLOv12, have played increasingly significant roles due to their compromise between speed and accuracy. 
% However, there is a common shortcoming of multi-scale feature fusion for YOLO series algorithms.

\begin{figure}[htb]
\centering
\includegraphics[width=1\linewidth]{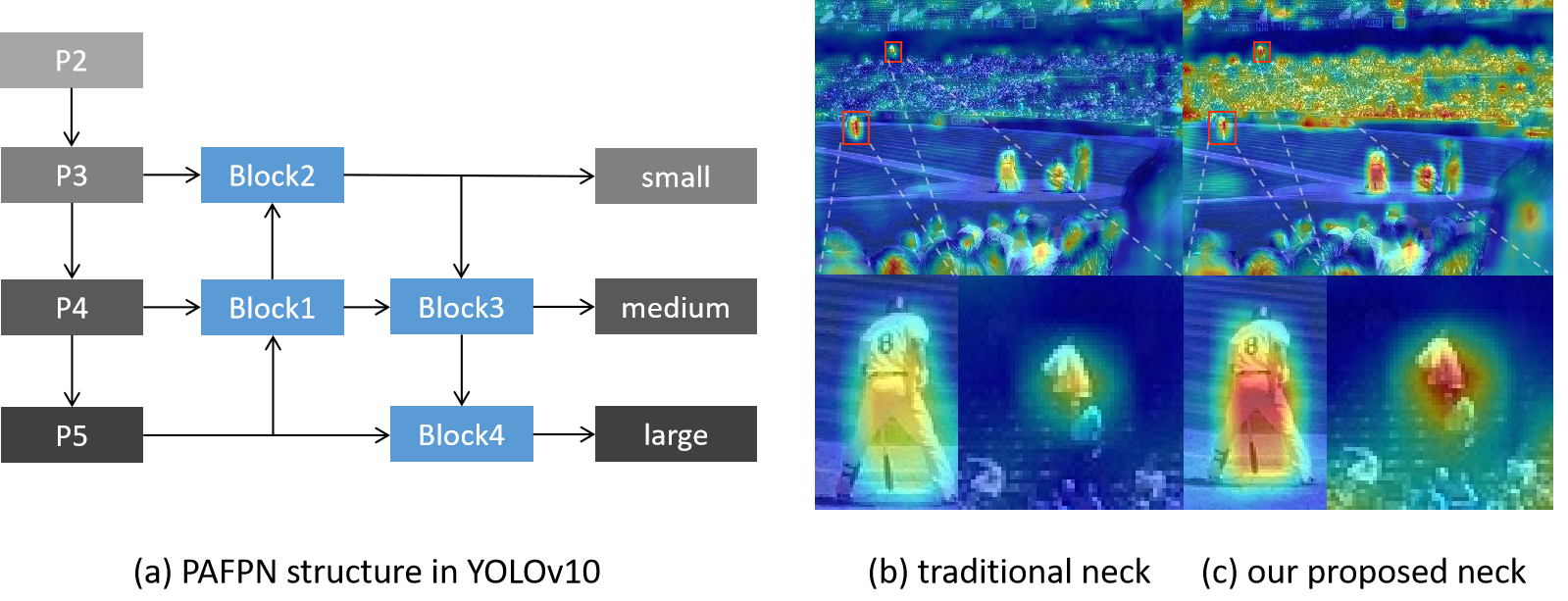}
\caption{Diagram (a) illustrates the PAFPN structure in YOLOv10, while diagrams (b) and (c) display the GradCAM++ visualizations of the feature map results from the same output head for the traditional PAFPN structure and the proposed MAFPN, respectively.}
\label{fig1}
\end{figure}

% The Feature Pyramid Network (FPN)~\cite{fpn} employed a top-down structure that enhances low-level features with semantic information from higher levels, thereby producing multi-scale feature maps. 
The Feature Pyramid Network (FPN)~\cite{fpn} utilizes a top-down architecture to enrich low-level features with high-level semantic information, effectively generating multi-scale feature maps.
Building on FPN, Path Aggregation Feature Pyramid Network (PAFPN)~\cite{pafpn} introduces a bottom-up path, allowing precise localization information from lower layers to be more effectively transmitted upwards. This enhancement improves the overall localization capabilities of the feature pyramid. Moreover, because of its straightforward and efficient fusion mechanism, PAFPN has been extensively employed in the YOLO series of models. In Fig.~\ref{fig1}(a), the layers P2-P5 represent the output information of different levels of the backbone. The neck structure of the YOLO series utilizes a traditional PAFPN, which incorporates two main paths for multi-scale feature fusion. Nevertheless, we found that PAFPN still possesses two significant limitations.

% However, a common shortcoming of YOLO series algorithms is the limitation of multi-scale feature fusion. 
% Although the feature fusion mechanism of the Path Aggregation Feature Pyramid Network (PAFPN)~\cite{pafpn}, an improvement over the Feature Pyramid Network (FPN)~\cite{fpn}, has been widely integrated into YOLO. 

% This mechanism introduces a dual-path approach to enhance feature integration, thereby improving accuracy while also controlling computational costs.

Firstly, the PAFPN structure primarily focuses on merging feature maps of similar scales but falls short in effectively processing and integrating multi-scale information from different resolution layers. This conservative approach to feature fusion may hinder the model's ability to fully engage across various levels, potentially leading to the loss of detailed information in deeper layers and producing overly simplistic results at each scale. For example, in Block1 of PAFPN, the input merges the up-sampled P5 layer with the adjacent P4 layer, neglecting vital shallow, low-level spatial details present in the P3 layer. Similarly, in Block2, there is a notable absence of direct fusion with the P2 layer, which is essential for capturing small target details. This shortfall is also evident in Block3 and Block4, limiting the overall effectiveness of the feature fusion process.
% (1). PAFPN tends to merge homogeneous scale feature maps but lacks integrated processing and fusion of multi-scale information from different resolution layers. This feature fusion approach may be overly conservative, preventing the model from fully interacting across different levels. As a result, it can lead to information loss in deeper layers and overly simplistic outputs at each level. For instance, in Block1 of PAFPN, the input consists of the up-sampled P5 layer and the sibling P4 layer, which overlooks the importance of shallow low-level spatial information in the P3 layer. Similarly, in Block2, there is no direct fusion of the P2 layer, which contains crucial information about small targets. This deficiency persists in Block3 and Block4 as well.
% Secondly, the architecture strategy for the small target detection layer is formulated through a singular down-top pathway and two relatived blocks, significantly impairing the model's proficiency in learning and representing small object features.

Secondly, the architectural strategy for the small target detection layer is designed using a single top-down pathway and two related blocks. This configuration significantly diminishes the model's ability to effectively learn and represent small object features, as the small target detection layer lacks sufficient supplementary information from additional feature layers.
What is more, each feature extraction module in PAFPN typically consists of an improved Cross Stage Partial Network (CSPNet)~\cite{cspnet} and fixed $3 \times 3$ convolutions, which restricts the network’s flexibility and limits its ability to capture larger receptive fields. 
In practical applications, these limitations may cause PAFPN to perform poorly in scenarios where objects of varying scales are distributed simultaneously or in dense small object scenarios. For example, in Fig.~\ref{fig1}(b) and (c), the YOLOv10 model with PAFPN shows significantly lower activation levels for dense small crowds compared to the MAFPN proposed in this paper.

\begin{figure}[htbp]  
    \centering  
    \begin{minipage}[b]{0.48\textwidth}  
        \centering  
        \includegraphics[width=\textwidth]{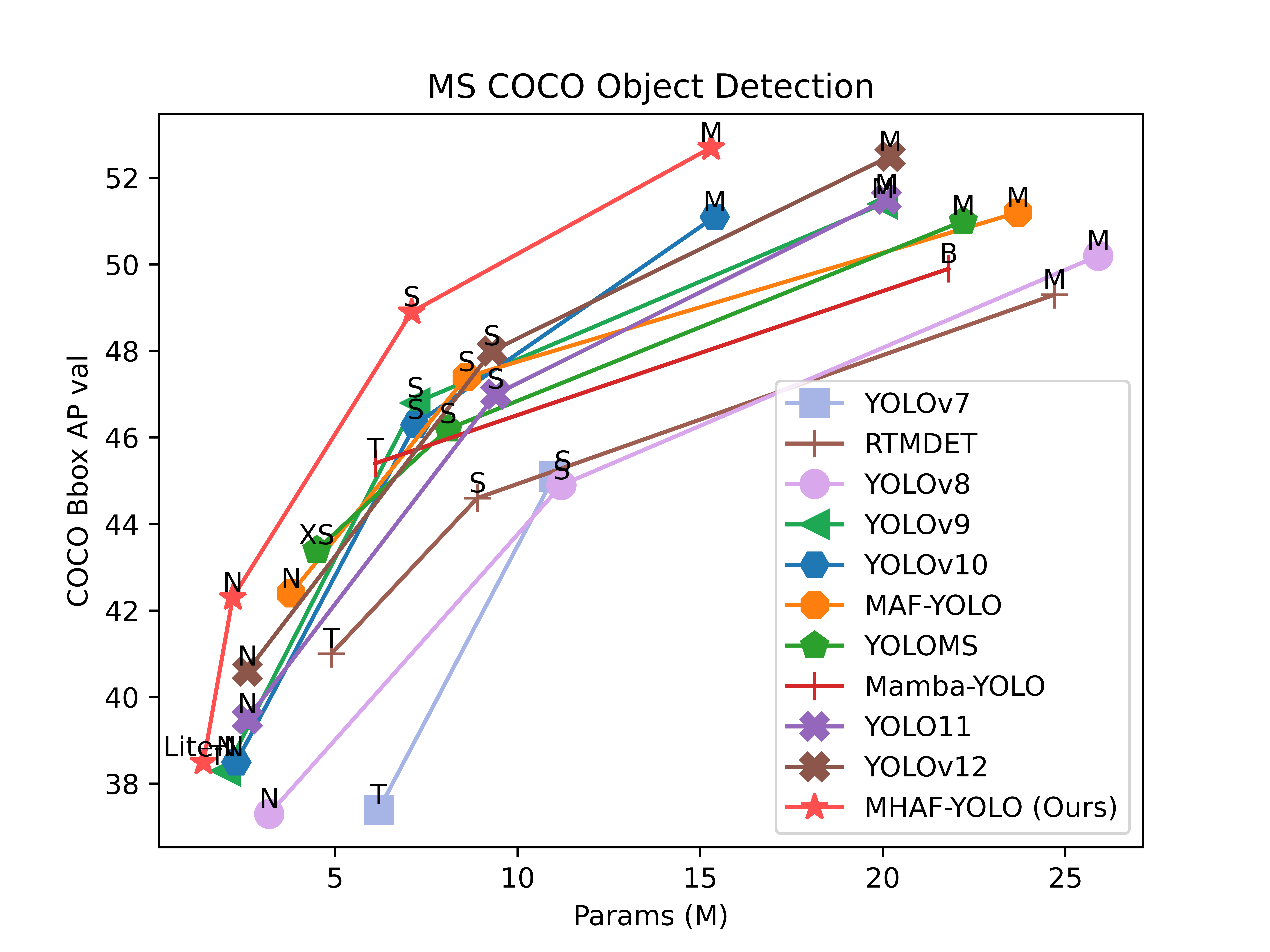}  
        % \footnotesize{(a)}  
    \end{minipage}  
    \hfill  
    \begin{minipage}[b]{0.48\textwidth}  
        \centering  
        \includegraphics[width=\textwidth]{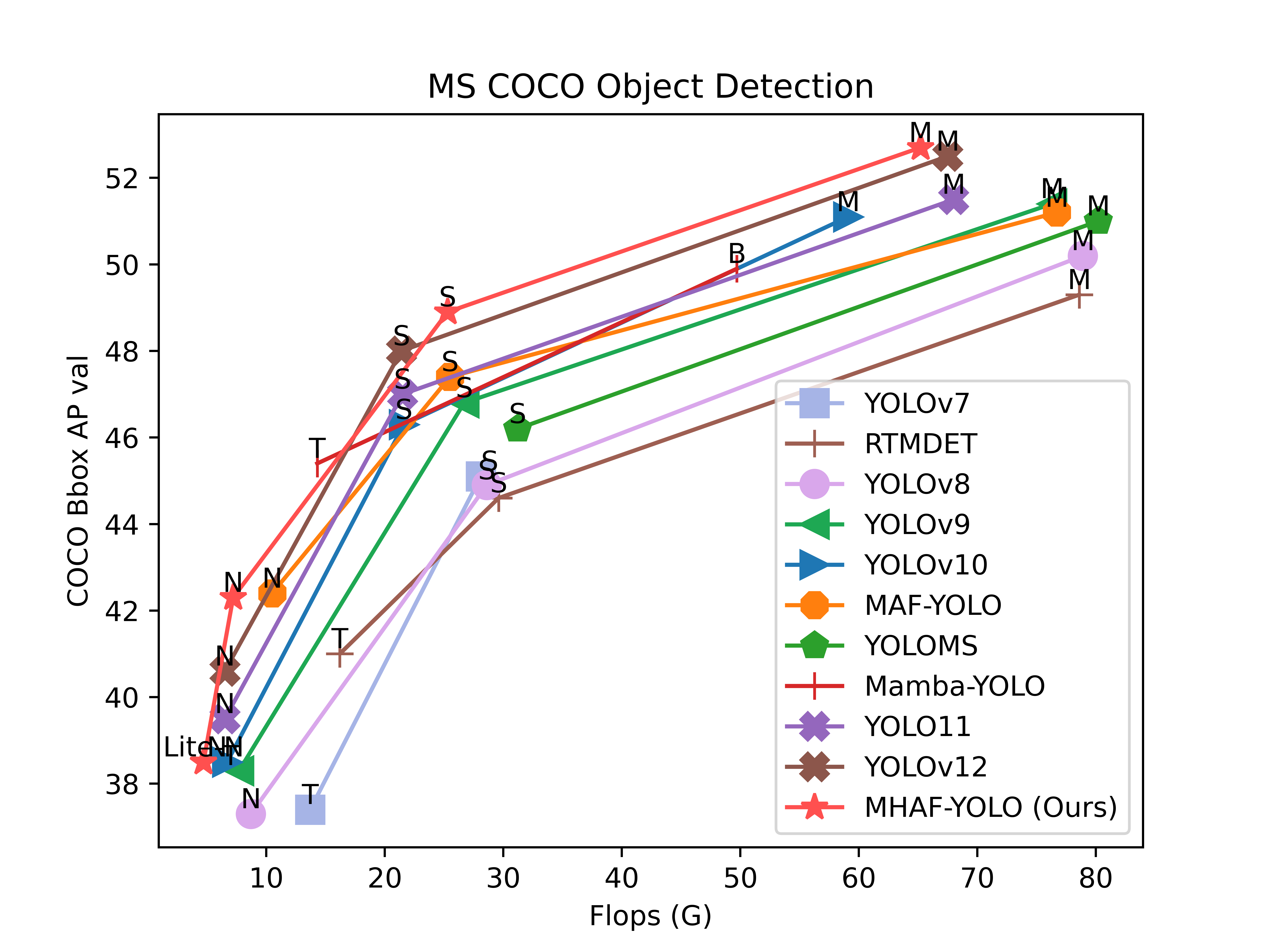}  
        % \footnotesize{(b)}  
    \end{minipage}   
    \caption{Comparisons with other state-of-the-art real-time object detectors in terms of parameter-accuracy (left) and GFlops-accuracy (right) trade-offs.}  
    \label{fig2}  
\end{figure}

We conducted extensive experiments to validate the effectiveness of MHAF-YOLO, scaling the model size to offer lite-nano, nano, small, and medium variants for diverse application scenarios. As illustrated in Fig.~\ref{fig2}, MHAF-YOLO achieves the highest accuracy with fewer parameters and lower computational costs, surpassing all state-of-the-art (SOTA) YOLO detectors. And the reduced computational load is particularly valuable on devices with limited computing resources. The main contributions of this paper are summarized as follows:

\begin{itemize}
\item[$\bullet$] We propose a new plug-and-play neck called Multi-Branch Auxiliary FPN (MAFPN) to achieve richer feature interaction and fusion. In MAFPN, Superficial Assisted Fusion (SAF) maintains shallow backbone information via bi-directional connectivity, enhancing the network's ability to detect small targets. Additionally, Advanced Assisted Fusion (AAF) enriches the gradient information of the output layer through multi-directional connections. Furthermore, MAFPN can be seamlessly integrated into any other detector to enhance its multi-scale representation capability. 
\item[$\bullet$] We designed the Reparameterized Heterogeneous Multi-Scale (RepHMS) module, featuring a high parameter utilization rate. This module expands the scope of perception by parallelizing a large kernel convolution with several small kernel convolutions without incurring additional inference costs while preserving information about small objects. RepHMS can be seamlessly integrated into either the backbone or FPN, enhancing the performance of any network.
\item[$\bullet$] We propose a Global Heterogeneous Flexible Kernel Selection (GHFKS) mechanism, which adaptively enlarges the effective receptive field of the entire network by adjusting the kernel sizes in RepHMS across different resolution feature layers of the network.
\item[$\bullet$] The Multi-Branch Heterogeneous Auxiliary Fusion YOLO (MHAF-YOLO) with its extremely high parameter utilization rate, achieves state-of-the-art performance in object detection of the COCO dataset, surpassing existing real-time object detectors. Furthermore, MHAF-YOLO has shown superior performance in instance segmentation and rotated object detection, demonstrating its strong generalization ability.
\end{itemize}

\section{Related Work}
\subsection{Real-time object detectors}
In practical applications, to meet the numerous detection tasks, faster and stronger networks are more widely used. Real-time object detectors have evolved from this trend, become an integral part of many applications that require instant analysis and interaction with dynamic environments. YOLOv1-YOLOv3~\cite{yolo1,yolo9000,yolov3} pioneer one-stage real-time detection and lay the foundation for a multi-scale detection framework. YOLOv4~\cite{yolov4} introduces the PAFPN structure and also utilizes mosaic and mixup data augmentation, which have been used ever since. YOLOv5~\cite{yolov5} designs a more optimized one-to-many label assignment based on anchor boxes, improves data augmentation techniques, and consolidates a CNN-based real-time general object detection framework that includes a backbone, multi-scale feature pyramid, and dense detection heads. YOLOX~\cite{yolox} first proposes the YOLO detection framework based on the dynamic label assignment strategy, removing the anchor boxes that rely on artificial priors or the inductive bias of the data itself. This achieves completely anchor-free multi-scale label assignment. 
Following that, PPYOLOE~\cite{ppyoloe} and YOLOv6~\cite{yolov6} explored reparameterization techniques and adopted the Task Alignment Learning (TAL)~\cite{tal} strategy for label assignment. The reparameterization technique enhances the representational capacity of individual convolutions without increasing the network’s parameter count or inference speed, thus providing a new direction for optimizing subsequent YOLO models. Meanwhile, the efficient performance of the TAL method has standardized the label assignment strategy across these models.
% Following that, PPYOLOE~\cite{ppyoloe} and YOLOv6~\cite{yolov6} explore reparameterization techniques and adopted the Task Alignment Learning (TAL)~\cite{tal} strategy in label assignment, the reparameterization technique enriches the representational capacity of individual convolutions without increasing the network’s parameter count or inference speed, providing a new direction for the optimization of subsequent YOLO models, and the efficient performance of the TAL method has unified the label assignment strategy for subsequent YOLO models. 
YOLOv7~\cite{yolov7} proposes the Efficient Layer Aggregation Network (ELAN) scheme to optimize the Cross Stage Partial Network structure from YOLOv4 and designs several trainable bag-of-freebies methods for lossless model optimization, making a contribution to the lightweight YOLO network. 
% Gold-YOLO~\cite{goldyolo} introduces an advanced Gather-and-Distribute (GD) mechanism, implemented using convolution and self-attention operations, to better integrate and distribute information across different layers, thereby enhancing feature fusion capabilities. However, this comes at the cost of a significantly increased parameter overhead.
DAMO-YOLO~\cite{damo} uses MAE-NAS~\cite{maedet} to search the backbone network under the constraints of low latency and high performance, and combines knowledge distillation to further improve the performance of the detector. 
YOLOv8~\cite{yolov8} fully absorbs the widely proven effective dynamic label allocation strategy, anchor-free detection architecture and ELAN module design concept, and extends the YOLO architecture to the field of segmentation and posture estimation. YOLOv9~\cite{yolov9} explores the issue of information bottleneck occurring when data passes through deep networks and introduces the concept of Programmable Gradient Information (PGI) to handle the various changes required for deep networks to achieve multiple goals. Additionally, it further realizes the lightweighting of the network architecture. YOLOv10~\cite{yolov10} uses one-to-many and one-to-one detection heads to effectively push the YOLO framework to an end-to-end reasoning paradigm, eliminating the post-processing step of non-maximum suppression. YOLO11~\cite{yolo11} represents the latest advancement in YOLO detection technology, introducing feature extraction modules called C3k2 and C2PSA, which significantly enhance the performance of each model. 

Throughout the evolution of YOLO models, most versions have focused on optimizing and improving their foundational convolutional modules. However, relatively few YOLO models have addressed the challenges of feature fusion in depth. Notably, even the latest versions, YOLOv7 to YOLO11, still rely on the original PAFPN structure. This paper, therefore, conducts a more comprehensive investigation into both the popular direction of convolutional module optimization and the less-explored area of feature fusion enhancement.

\subsection{Multi-scale features fusion for object detection}
The different sizes of targets are a major characteristic of detection tasks. Feature maps of different scales correspond to object information of different sizes. Usually, low-dimensional information is used to represent small objects, and high-dimensional information is used to represent large objects. There is an implicit correlation between the features of each dimension. As the network depth increases, the low-level texture features will be transformed into high-level semantic information. How to enhance the connection between features of different levels is the focus of many works. 

Feature Pyramid Networks (FPN) is the first algorithm to propose feature fusion in object detection. The original intention of FPN is to enhance the multi-scale detection capability of networks by incorporating cross-scale connections and information exchange. However, the bottom-up fusion method of FPN is slightly simpler. It only transfers the semantic information of the high-level layer to the low-level layer, but the texture information of the low-level layer is not transferred to the high-level layer. The Path Aggregation Network (PANet) adopts a bottom-up path, which makes the information fusion between different levels more adequate based on FPN. YOLOv6 uses a bidirectional connection (BIC) mechanism to better utilize the shallow information of the backbone, and the processing of high-resolution features and low-resolution features can take into account both large and small target detection. Furthermore, Asymptotic Feature Pyramid Network (AFPN)~\cite{afpn} starts by fusing two adjacent Low-Level features and gradually incorporates High-Level features into the fusion process. In this way, large semantic gaps between non-adjacent Levels can be avoided. DAMO-YOLO adopts the Reparameterized Generalized FPN (RepGFPN) to achieve a richer fusion of the backbone and neck. Gold-YOLO~\cite{goldyolo} proposes a Gather-and-Distribute mechanism (GD), which is implemented through convolution and self-attention, further improving the fusion capability of multi-scale features. Information of each scale is collected and fused through a unified module, and then the fused features are distributed to different layers. This not only avoids the inherent information loss of the traditional FPN structure, but also enhances the information fusion capability of the Neck part without significantly increasing the latency. These methods comprehensively consider the characteristics of multi-scale features and contribute to feature fusion in object detection, but there is still room for further optimization.

\subsection{Foundational Convolutional Module}
\begin{figure}[htbp]  
    \centering  
    \begin{minipage}[b]{0.3\textwidth}  
        \centering  
        \includegraphics[width=\textwidth]{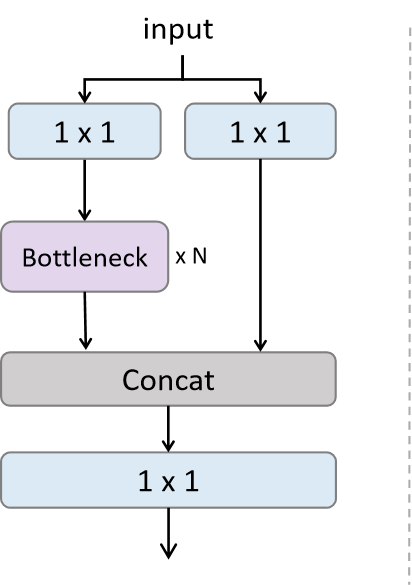}  
        \footnotesize{(a) 
        % C3/CSPRes/BepC3/CSPNextBlock
}  
    \end{minipage}  
    \hfill  
    \begin{minipage}[b]{0.29\textwidth}  
        \centering  
        \includegraphics[width=\textwidth]{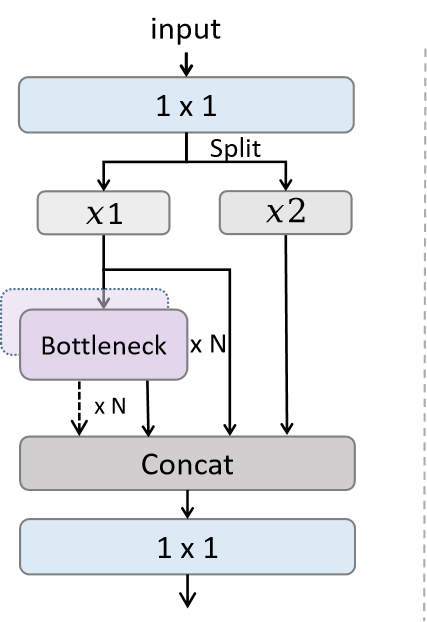}  
        \footnotesize{(b) 
        % ELAN/C2f/GELAN/RepHELAN/C2CIB
}  
    \end{minipage}  
    \hfill  
    \begin{minipage}[b]{0.33\textwidth}  
        \centering  
        \includegraphics[width=\textwidth]{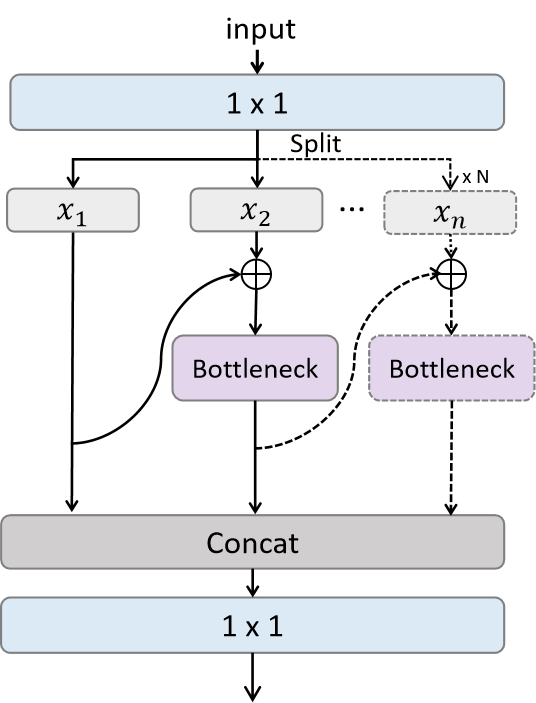}  
        \footnotesize{(c) \\
        % MSBlock    \\       
         }  
    \end{minipage}  
    \caption{Different feature extraction blocks. (a)The basic block of CSPNet in~\cite{yolov5, ppyoloe, yolov6, rtmdet}. (b)The proposed block with ELAN in~\cite{yolov7, yolov8, yolov9, mafyolo, yolov10, yolo11}. (c)The basic unit of YOLOMS~\cite{yoloms}.}  
    \label{fig3}  
\end{figure}
CSPNet~\cite{cspnet} alleviates the issue that previous works required heavy inference computations from the network architecture perspective. It effectively improves accuracy while reducing a large amount of computations, and it also possesses strong ease of use and versatility. In most early YOLO networks, the CSPNet structure was widely used as the basic feature extraction module. The latest state-of-the-art YOLO models employ variants of the CSPNet to achieve superior performance. These modules are integrated into the Backbone and Neck of the network. Notably, while these modules share a similar overall architecture, their bottleneck structures of differ slightly. Different bottleneck consist of several regular convolutions, DW convolutions, reparameterized convolutions, and so on.

As shown in Fig.~\ref{fig3}(a), the C3 module in YOLOv5, the CSPRes module in PPYOLOE, the BepC3 module in YOLOv6, and the CSPNextBlock in RTMDet are all classic CSPNet structures. The input is divided into two branches by a $1 \times 1$ convolution, each branch having half channels of the original input. The first branch undergoes deep feature extraction through multiple bottleneck and then concatenated with the second branch before the final output.

In designing its network architecture, YOLOv7 introduces considerations for gradient propagation efficiency to balance the network's learning capability. Compared to CSPNet, the design of ELAN places greater emphasis on optimizing the gradient pathways, reducing issues related to gradient vanishing and exploding. This enhances the stability and convergence speed of the model during training. Moreover, ELAN maintains high detection accuracy while further reducing unnecessary computational load and the number of parameters, which in turn increases the inference speed of the model. It develops the ELAN, as shown in Fig.~\ref{fig3}(b), which is used in later models such as the C2f module in YOLOv8, the GELAN module in YOLOv9, the RepHELAN module in MAF-YOLO, the CIB module in YOLOv10, and the C3k2 module in YOLO11. The ELAN variant replaces the two $1 \times 1$ convolutions in the CSPNet with a split operation and retains the output of each bottleneck. These outputs are concatenated before the final output.

The MSBlock proposed in YOLOMS offers significant enhancements in multi-scale feature representation compared to CSPNet. It utilizes a hierarchical feature fusion strategy, employing multiple branches for feature extraction and introducing convolutional kernels of varying sizes into the backbone network to capture features at different scales. This design enables the model to handle objects of diverse sizes more effectively, thereby improving detection accuracy. Furthermore, unlike the standard convolutions or re-parameterized convolutions used in previous YOLO versions, MSBlock incorporates an Inverted Bottleneck Block combined with Depthwise Convolution to reduce computational costs. As illustrated in Fig.~\ref{fig3}(c), the MSBlock enhances the network's depth by employing width augmentation and cascade connections. The input is divided into N branches, each processed through a bottleneck module. The output from each branch is then passed to the subsequent branch, enabling efficient transmission of deep feature information throughout the network.

\section{Method}
\subsection{Macroscopic architecture}
\begin{figure}[htb]
	
	\begin{minipage}[b]{1.0\linewidth}
		\centering
		\centerline{\includegraphics[width=13cm]{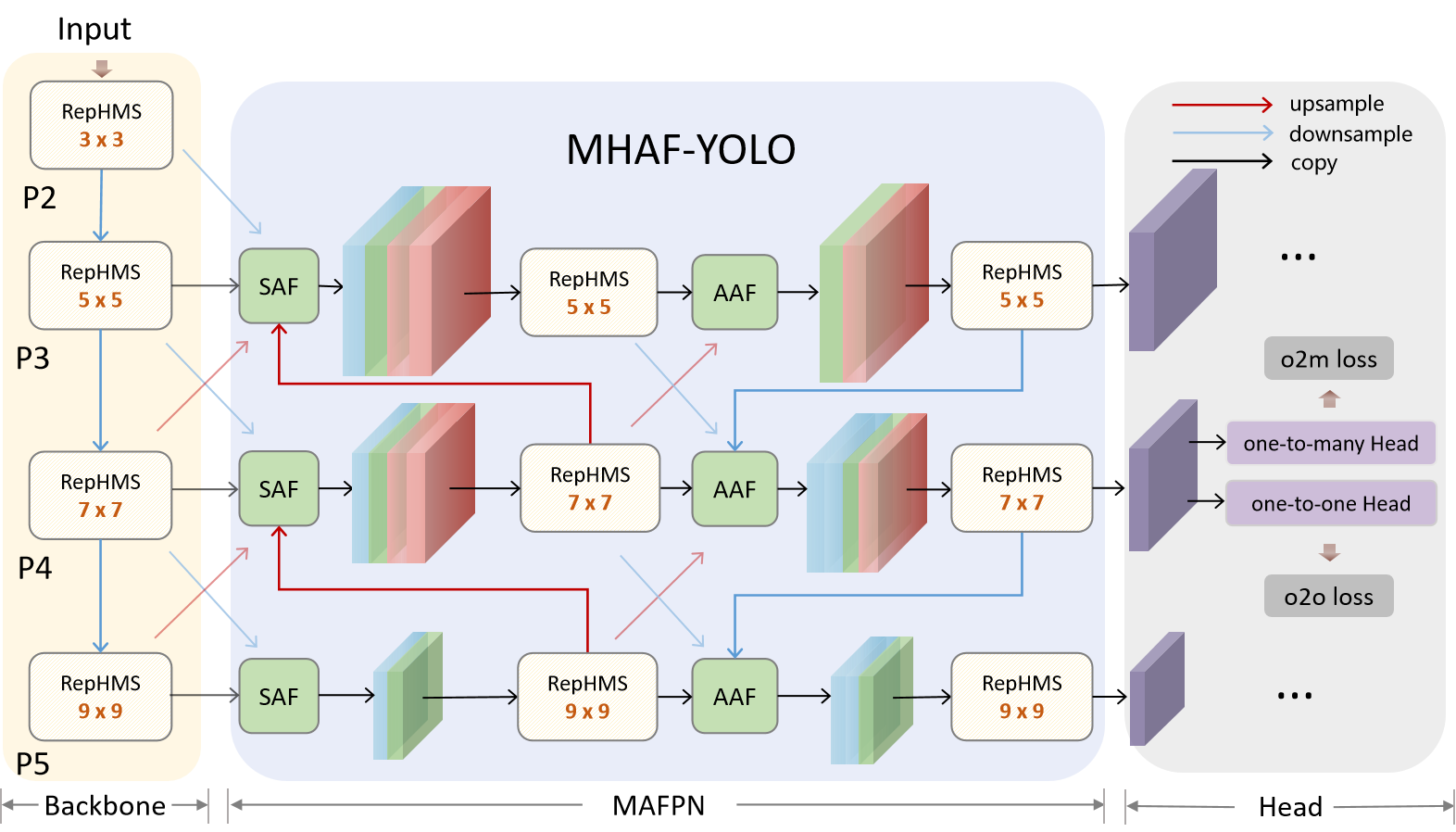}}
		%  \vspace{2.0cm}
		% \centerline{(a) Result 1}\medskip
	\end{minipage}
	\caption{Overview of the network architecture of MHAF-YOLO.}
	\label{mayolo}
\end{figure}

As illustrated in Fig.~\ref{mayolo}, we break down the macroscopic architecture of one-stage object detector into three main components: backbone, neck, and head. Within the proposed MHAF-YOLO, the input image initially passes through the backbone, which consists of four stages: P2, P3, P4, and P5. MAFPN is designed as a neck structure. In the first bottom-up pathway, the SAF module is responsible for extracting multi-scale features from the backbone and performing preliminary assisted fusion at the shallow layers of the neck. Meanwhile, AAF collects gradient information from each deep layer through dense connections in the second top-down pathway, ultimately guiding the head to obtain diversified output information across three resolutions. Both of the aforementioned structures employ the RepHMS module for feature extraction, seamlessly integrating the GHFKS concept to utilize dynamically sized convolutional kernels and achieve adaptive receptive fields across different layers. Finally, the detection heads predict object bounding boxes and their corresponding categories based on feature maps at each scale to compute their loss.

\subsection{Global Adaptive Heterogeneous Flexible Kernel Selection mechanism}
An important factor contributing to the effectiveness of transformers is their self-attention mechanism, which performs query-key-value operations over a global or larger window scale. Similarly, large convolutional kernels capture both local and global features, and the use of moderately large convolutional kernels to increase the effective receptive field has been demonstrated in several works to be effective. Research conducted by Trident Network~\cite{trident} suggests that networks with larger receptive fields are preferable for detecting larger objects, while inversely, smaller-scale targets benefit from smaller receptive fields. YOLOMS~\cite{yoloms} introduced the concept of Heterogeneous Kernel Selection (HKS) protocols. Employing an incremental convolutional kernel design of 3, 5, 7, and 9 in the backbone to balance performance and speed. Inspired by this work, we extend it to the Global Heterogeneous Flexible Kernel Selection (GHFKS) mechanism, integrating the concept of heterogeneous large convolutional kernels throughout the entire MHAF-YOLO architecture. In addition to the progressively increasing convolutional kernels in RepHMS of the backbone, we also introduce large convolutional kernels of 5, 7, and 9 in MAFPN to adapt to the requirements of different resolutions, thus progressively obtaining multi-scale sensory field information.

\subsection{Multi-Branch Auxiliary FPN}
Accurate localization relies on detailed edge information from shallow networks, while precise classification requires deeper networks to capture coarse-grained information~\cite{yolov3}. We believe that an effective FPN should support full and sufficient convergence of shallow and deep network information flows.

\subsubsection{Superficial Assisted Fusion.}
\begin{figure}[htb]
	
	\begin{minipage}[b]{1.0\linewidth}
		\centering
		\centerline{\includegraphics[width=7cm]{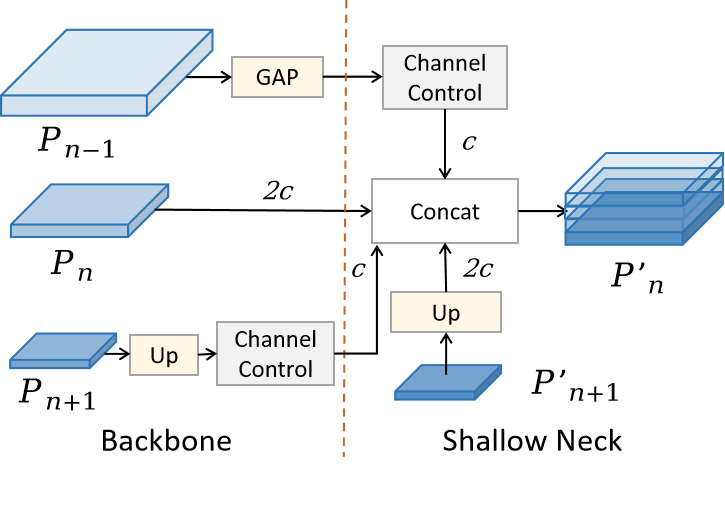}}
		%  \vspace{2.0cm}
		% \centerline{(a) Result 1}\medskip
	\end{minipage}
	\caption{The architecture of Superficial Assisted Fusion.}
	\label{fig5}
\end{figure}
Preserving shallow spatial information in the backbone is crucial for enhancing the detection capability of smaller objects. However, the information supplied by the backbone is relatively elementary and prone to interference. Therefore, we incorporate shallow information as assisted branches into the deeper network to ensure the stability of subsequent layer learning. Following these principles, we have developed the SAF module, delineated in Fig.~\ref{fig5}. 
The primary objective of SAF is to integrate deep-level information with the shallow spatial information embedded in multi-scale feature layers within the backbone, aiming to preserve abundant localization details to enhance the spatial representation of the network. Additionally, we utilize $1 \times 1$ convolutions to control the number of channels in shallow layer information, ensuring it occupies a smaller proportion during the concat operation without affecting subsequent learning. Let $P_{n-1}$, $P_n$ and $P_{n+1} \in R^ {H \times W \times C}$ represent the feature maps at different resolutions, where $P_{n}$, $P^{\prime}_{n}$ and $P^{\prime\prime}_{n} $ denote the feature layers of the backbone, and the two paths of the MAFPN. The notation $ U(\cdot) $ signifies the up-sampling operation. $GAP$ stands for Global Average Pooling, $Down$ denotes a $ 3 \times 3$ downsampling convolution accompanied by a batch normalization layer, and $ \delta $ represents a $ silu $ function, $Conv$ represents the $ 1 \times 1$ convolution of the number of control channels. The output result after applying SAF is as follows:
\begin{equation}
  P^{\prime}_n = concat(\delta(GAP(P_{n-1})), P_{n},\delta(Conv(U(P_{n+1}))), U(P^{\prime}_{n+1}))     
\end{equation}

\begin{figure}[htb]
	
	\begin{minipage}[b]{1.0\linewidth}
		\centering
		\centerline{\includegraphics[width=7cm]{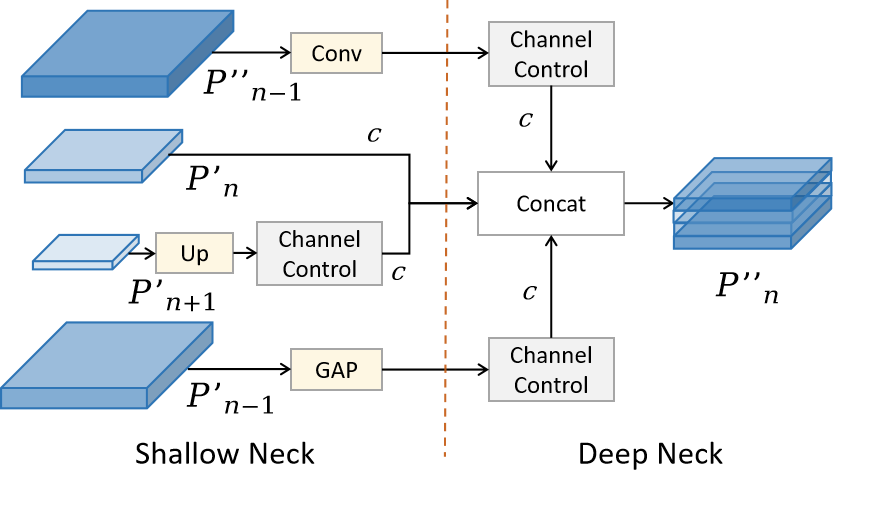}}
		%  \vspace{2.0cm}
		% \centerline{(a) Result 1}\medskip
	\end{minipage}
	\caption{The architecture of Advanced Assisted Fusion.}
	\label{AAF}
\end{figure}

\subsubsection{Advanced Assisted Fusion.}
To further enhance the interactive utilization of feature layer information, we employ the AAF module in the deeper layers of the MAFPN for multi-scale information integration.
Specifically, Fig.~\ref{AAF} illustrates the AAF connections in $P^{\prime\prime}_{n}$, which involve information aggregation across the shallow high-resolution layer $P^{\prime}_{n+1}$, the shallow low-resolution layer $P^{\prime}_{n-1}$, the sibling shallow layer $P^{\prime}_{n}$, and the previous layer $P^{\prime\prime}_{n-1}$.
At this moment, the final output layer P4 can merge information from four different layers simultaneously, thereby significantly enhancing the performance of medium-sized targets. AAF also employs $ 1 \times 1$ convolutional control channels to regulate the impact of each layer on the outcome. Through experimentation, we found that when the strategy in SAF is used, i.e., the number of channels in the three shallow layers is set to half of the number of channels in the deeper layers, which in turn results in a slight degradation of performance. Drawing from the conventional single-path architecture of the FPN, we postulate that the initial guiding information is already embedded within the shallow layers of the MAFPN. Consequently, we equalize the number of channels across each layer to ensure the model obtains diverse outputs. The output result after applying AAF is as follows:
\begin{equation}
  P^{\prime\prime}_n = concat(\delta(Down(P^{\prime}_{n-1})), \delta(GAP(\!P^{\prime\prime}_{n-1})), P^{\prime}_{n}, \delta(Conv(U(P^{\prime}_{n+1}))))    
\end{equation}

\subsection{Re-parameterized Heterogeneous Multi-Scale Module}
\begin{figure}[htb]
	
	\begin{minipage}[b]{1.0\linewidth}
		\centering
		\centerline{\includegraphics[width=13cm]{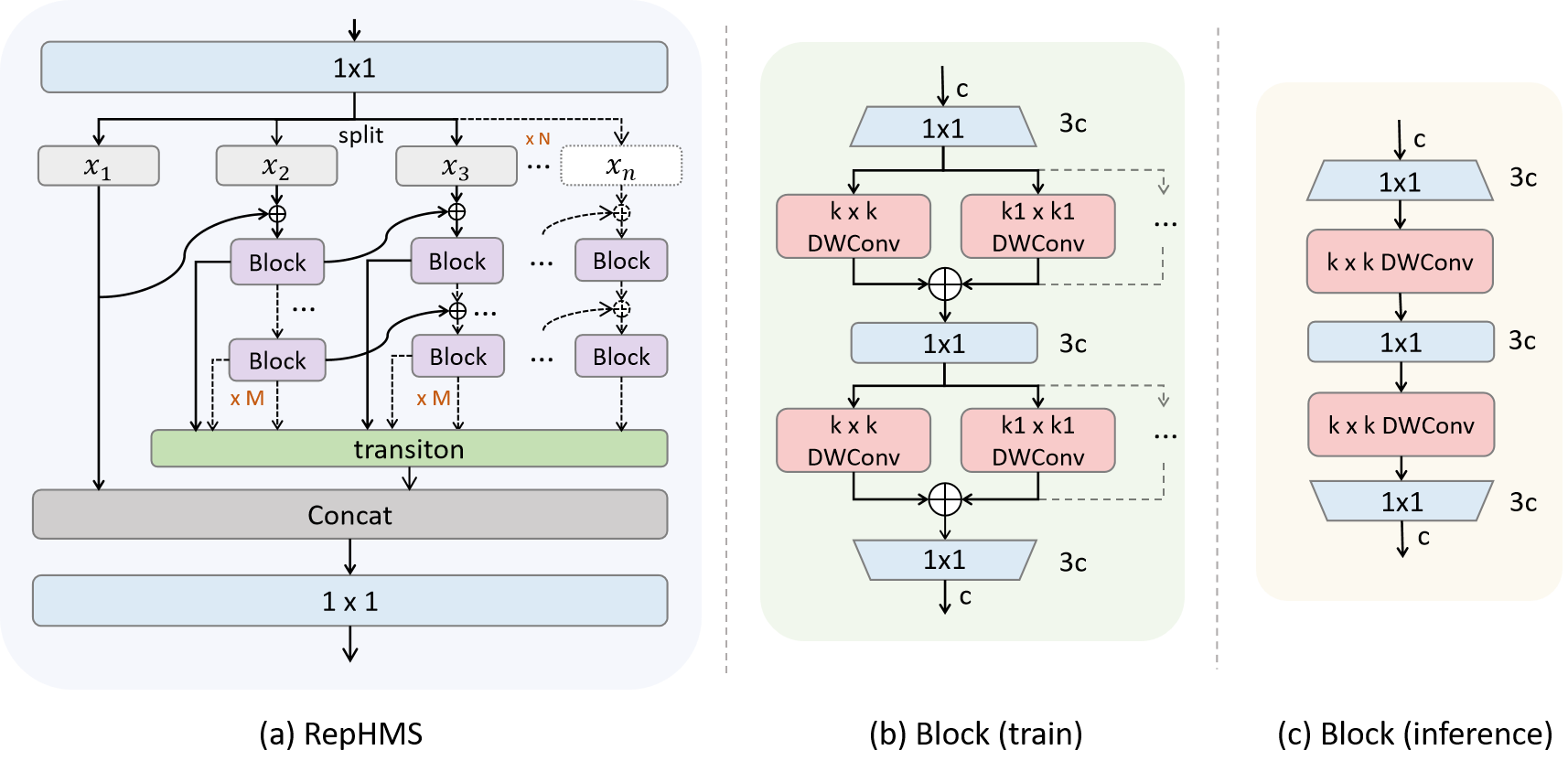}}
		%  \vspace{2.0cm}
		% \centerline{(a) Result 1}\medskip
	\end{minipage}
	\caption{The architecture of RepHMS.}
	\label{fig7}
\end{figure}
% \begin{figure}[htbp]  
%     \centering  
%     \begin{minipage}[b]{0.45\textwidth}  
%         \centering  
%         \includegraphics[width=\textwidth]{RepHMS.png}  
%         \footnotesize{(a) RepHMS
% }  
%     \end{minipage}  
%     \hfill  
%     \begin{minipage}[b]{0.28\textwidth}  
%         \centering  
%         \includegraphics[width=\textwidth]{RepHMS-1.png}  
%         \footnotesize{(b) Block (train)

% }  
%     \end{minipage}  
%     \hfill  
%     \begin{minipage}[b]{0.24\textwidth}  
%         \centering  
%         \includegraphics[width=\textwidth]{RepHMS-2.png}  
%         \footnotesize{(c) Block (inference)     
%          }  
%     \end{minipage}  
%     \caption{123}  
%     \label{fig:example}  
% \end{figure}
After designing the MAFPN structure in the preceding section, another challenge lies in efficiently designing the feature extraction block within the entire architecture. This section presents the design of a powerful encoder architecture that efficiently learns expressive multi-scale feature representations, boasting an extremely high parameter utilization rate. The structure of RepHMS is shown in Fig.~\ref{fig7}(a). Initially, the input information undergoes a $1 \times 1$ convolution and a Split operation, producing N information streams. The first branch retains the original shallow information. Starting from the second branch, the input information passes through M concatenated blocks to enhance feature extraction capabilities. Incorporating the idea of ELAN, the output of each block is retained and integrated into the final output layer. Additionally, each branch includes the cascade concept, allowing even parallel branches to receive shallow information from the previous branch, thus enriching the gradient flow. The last branch outputs the deepest level of information, and the final concatenation and $1 \times 1$ convolution operations integrate and output the diverse branch information. By adjusting the coefficients M and N, we can easily control the feature extraction capability of RepHMS. RepHMS preserves the gradient flow information in each branch as much as possible, while progressively integrating deeper information from the previous layer through cascade connections. As the process continues, the information in the branches becomes increasingly diverse, and feature extraction becomes more thorough, optimizing the representation of information to its fullest extent. As a result, the RepHMS module can be seamlessly integrated into any advanced detector, significantly enhancing its performance.

As shown in Fig.~\ref{fig7}(b), each block consists of several DW convolutions, combined with advanced re-parameterization techniques to achieve high parameter efficiency. The first $1 \times 1$ convolution expands the channels, and each RepHConv is followed by a point convolution to compensate for the performance loss of the DW convolution. The final convolution scales the number of channels.

% Initially, the input information undergoes a $1 \times 1$ convolution and a Split operation, resulting in two streams. One stream preserves the original information, which then directly enters the Concat operation, while the other stream undergoes processing through N Inverted Bottleneck units. Due to the mechanism of ELAN, the branches and the outputs passing through each Inverted Bottleneck are retained and eventually concatenated together. The specific structure of the Inverted Bottleneck is illustrated in Fig.~\ref{RepHELAN}(b), where the input sequentially passes through a $ 1 \times 1$ convolution to expand the number of channels, followed by a $ k \times k$ RepHDWConv operation, and finally by $ 1 \times 1$ point convolution to shrink the number of channels and compensate for the possible loss of information caused by DWConv.
\subsection{Re-parameterized Heterogeneous Depthwise Convolution}
Firstly, we employed Depthwise convolution with a large kernel in the global architecture to implement the aforementioned GHFKS mechanism. Our study also indicates that while larger convolutional kernels may enhance performance by encoding more extensive regions, they might inadvertently obscure details relevant to small targets, thus leaving room for further improvement. Therefore, we transferred the heterogeneous idea from the global architecture to a single convolution and incorporated the idea of reparameterization~\cite{repvgg,replknet,replknetv2} to realize RepHConv. Specifically, we complement the detection of small targets by concurrently running large and small convolutional kernels. Different sizes of convolution kernels enhance both the network's ERF and the diverse representation of features. 
\begin{figure}[htb]
	
	\begin{minipage}[b]{1.0\linewidth}
		\centering
		\centerline{\includegraphics[width=13cm]{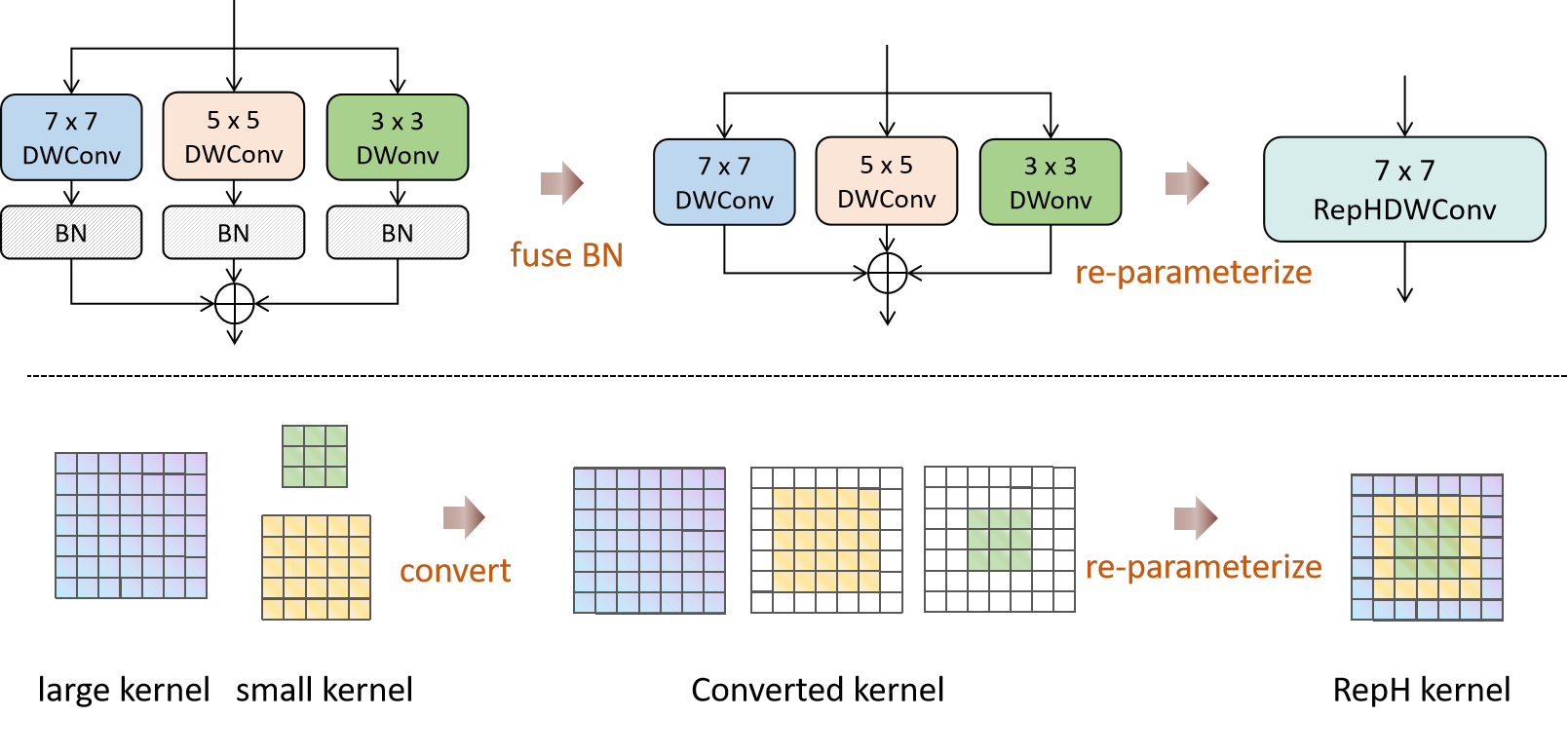}}
		%  \vspace{2.0cm}
		% \centerline{(a) Result 1}\medskip
	\end{minipage}
	\caption{An example of re-parameterizing a small kernel (e.g., $3 \times 3$, $5 \times 5$) into a large one (e.g., $7 \times 7$).}
	\label{fig8}
\end{figure}

As shown in Fig.~\ref{fig7}(b) and (c), the Block exhibits certain differences between training and inference. During training, the network runs n parallel depthwise convolutional (DWConv) operations of varying sizes, while during inference, these convolutions are merged into one, resulting in no decrease in inference speed. We believe that RepHDWConv is a superior convolutional strategy that enhances the representation capability across multiple scales with minimal loss.

Steps for the reparameterization of a $7 \times 7$ RepHDWConv is shown in Fig~\ref{fig8}.
Let $ \mu $, $ \sigma $, $\gamma$, $ \beta $ as the accumulated mean, standard deviation and learned scaling factor and bias
of the BN layer. $RepHDW(x)$ represents a RepHDWConv parameters. $I$ indicates input feature maps, $K_{n}$ and $B_{n}$ show the weight and bias of the convolution with a $n \times n$ kernel.

Firstly, a $k_1 \times k_1$ large DWConv and many $k_2 \times k_2$ small DWConv will be parallelized in a RepHDWConv, Each DWConv is followed by a batch normalization (bn) layer. Then the parameters of each convolution kernel will be merged with the parameters of its corresponding bn layer.
\begin{equation}
\begin{split}
  BN(RepHDW(x)) = \frac{\gamma   \cdot   (RepHDW(x) - \mu )}{{\sqrt{ \sigma ^2+ \epsilon } }} + \beta \\
 % = \frac{\gamma   \cdot   DW(x)}{{\sqrt{ \sigma ^2+ \epsilon } }} + (\frac{\gamma   \cdot (-\mu)}{{\sqrt{ \sigma ^2+ \epsilon } }} + \beta)
\end{split}
\end{equation}
Let $W_{fused}$ and $B_{fused}$ represent the parameters and bias of the convolution operation after BN fusion. Extracting the fused parameters $W_{fused}$ and $B_{fused}$ gives:
\begin{equation}
  W_{fused} = \frac{\gamma   \cdot   RepHDW}{{\sqrt{ \sigma ^2+ \epsilon } }}, B_{fused} = \frac{\gamma   \cdot (-\mu)}{{\sqrt{ \sigma ^2+ \epsilon } }} + \beta
\end{equation}
Then the Conv after fusing bn layers can be represented as:
\begin{equation}
  BN(RepHDW(x)) = W_{fused}(x) + B_{fused}
\end{equation}
In the second step, many small DWConv are equated to a large DWConv by means of padding, and then do re-parameterization. The parameters and bias of these heterogeneous DWConv are constructed by doing accumulation to get a new RepHDWConv, and the output feature map $O$ is:
\begin{equation}
	\begin{split}
O = I \otimes \left( K_{2n-1} + \sum_{i=1}^{m} K_{2n-(2i+1)} \right) \\+ \left( B_{2n-1} + \sum_{i=1}^{m} B_{2n-(2i+1)} \right)
	\end{split}
\end{equation}
$ \text{where } n \geq 3 \text{ and } m \text{ is the largest integer such that } 2n-(2m+1) \geq 3$. 

\section{Experiments}
\subsection{Experimental Setup}
\noindent
\textbf{Datasets.} To verify the effectiveness of the proposed method, we performed experiments on three authoritative public benchmark datasets, involving three different tasks: object detection, instance segmentation, and rotated object detection. The datasets used are as follows:
\begin{itemize}
\item[$\bullet$] \textbf{MS COCO} \cite{coco}. The COCO dataset, released by Microsoft in 2014, aims to provide a large and challenging multi-task image dataset. We evaluate the object detection task on this dataset using the train2017 set, which contains 118,287 images for training, and the val2017 set, which contains 5,000 images for validation. In addition, we also performed instance segmentation on the COCO dataset to assess the multi-task capability of our model.
% We use the train2017 set, which includes 118287 images for training, and the val2017, which includes 5000 images for validation.
% Furthermore, we have conducted tests on the COCO test2017 dataset, which comprises 20288 images, to validate the generalization capability of our model.
\item[$\bullet$] \textbf{Pascal VOC} \cite{voc}. We adhere to most of the mainstream VOC dataset configurations, which consist of 20 categories, utilizing the train2012, val2012, train2007, and val2007 datasets for model training, encompassing 16,551 images. The test2007 set, containing 4,952 images, is employed for validation and testing.
\item[$\bullet$] \textbf{DOTA-v1.0} \cite{dota}. DOTA-v1.0 is a dataset specifically designed for rotational object detection in remote sensing images, consisting of 2,806 images and containing 188,282 instances across 15 categories. We adopt the default configuration, using 1,411 images as training set, 458 images as the validation set, and 937 images as the test set.
% We adhere to most of the mainstream VOC dataset configurations, utilizing the train2012, val2012, train2007, and val2007 datasets for model training, which encompass 16,551 images. The test2007 set, containing 4,952 images, is employed for validation and testing.
% \item[$\bullet$] \textbf{VisDrone} \cite{visdrone}. This dataset contains 8,599 drone-captured images (6,471 for training, 548 for validation, and 1,580 for testing) with a resolution of about 2000 ×1500 pixels. These objects are from ten categories, with 540,000 instances annotated in the training set, most of which contain different categories of vehicles and pedestrians observed from drones, and the majority of which are small targets.

\end{itemize}
% \noindent
\subsection{Implementation Details}
\noindent
\textbf{Object detection and instance segmentation.} 
Our implementation is based on the YOLOv10 framework. For Object Detection task, All experiments are conducted with 4 NVIDIA GeForce RTX 2080Ti GPUs, and all the scales of MHAF-YOLO are trained from scratch for 500 epochs without relying on other large-scale datasets, like ImageNet~\cite{imagenet}, or pre-trained weights. We mostly followed the YOLOv10 settings and used the SGD optimizer to train, and we replaced the original mixup data augmentation strategies with the more advanced cached-mixup strategies from RTMDet~\cite{rtmdet} and used the lower probability copy-paste~\cite{copy} method. The last 10 epochs turn off these strong data enhancement strategies. 
For the instance segmentation task, we followed the configurations of YOLOv8, RTMDet, and YOLO11, only modifying the object detection head of MHAF-YOLO to an instance segmentation head to adapt to this task. We also used the same hyperparameter configuration and training epochs as in the object detection task, training the model from scratch.

\noindent
\textbf{Rotated object detection.} 
Firstly, we replaced the object detection head of MHAF-YOLO with the rotated object detection head from YOLO11 and substituted the IOU with rotated IOU to achieve this task. We unified the training process to 200 epochs and optionally applied multi-scale offline data augmentation, rotation augmentation, and vertical flipping. Additionally, to benchmark against other rotated object detection tasks, we trained the backbone of MHAF-YOLO for 300 epochs on the ImageNet dataset, which can also serve as a foundation for further fine-tuning.
 For single-scale training and testing, the original images are cropped into 1024×1024 patches. In the case of multi-scale training and testing, the original images are resized with scales of 0.5, 1.0, and 1.5, and then cropped into 1024×1024 patches. Each patch has an overlap of 500 pixels. For the evaluation metric, we utilize the same mAP calculation method as used in PASCAL VOC2007~\cite{voc}, with the difference that we employ rotated IoU to determine the matched objects.
% The detailed hyperparameters are shown in Tab.~\ref{hyperparameters}.

\subsection{Analysis of RepHMS}
In this subsection, we will perform a series of ablation studies on the RepHMS
module. By default, we use the MHAF-YOLO nano in all experiments.
\subsubsection{Exploring the flexibility of RepHMS}
We first do ablation experiments of RepHMS module with various computational blocks from other advanced YOLO models in Tab.~\ref{DifferentBlock}. All modules are used in both the Backbone and Neck of the network. To ensure the fairness of our experiments, we set similar depth and width coefficients so that the parameter count of each improved model is as close as possible. The RepHMS module, compared to other modules, boasts exceptional parameter efficiency by leveraging re-parameterized DWConv and multi-scale receptive fields, achieving an optimal balance between parameter count and accuracy.
% The RepHMS module, compared to other modules, boasts extremely high parameter efficiency, achieving a balance between parameter count and accuracy.

\begin{table}[H]
	\centering
	\caption{The impact of different computational blocks on the MHAF-YOLOn.}\label{DifferentBlock}
	\renewcommand\arraystretch{1.0}	
	\setlength{\tabcolsep}{5mm}
		\begin{tabular}{ccccc}
		\hline
		  Computational Blocks &  Param & FLOPs & $AP$ & $AP_{50}$ \\
    \hline
	     C2f~\cite{yolov8} &2.6M&7.5G&39.3 & 55.3\\
        MSBlock~\cite{yoloms} &2.2M&6.5G &41.2 & 57.6\\
        GELAN~\cite{yolov9} &2.1M&8.5G &40.3 & 55.7\\
       CIB~\cite{yolov10} &2.2M&6.2G &40.7 & 56.7\\
        RepHELAN~\cite{mafyolo} &2.4M&7.1G &41.7 & 58.1 \\
	    \rowcolor{gray! 8} RepHMS &2.2M&7.2G&42.3 & 58.5\\  
		\hline
		\end{tabular}
\end{table}

\subsubsection{Ablation study on RepHMS}
As seen in Tab.~\ref{ABRepHELAN}, we have performed an ablation study on the RepHMS module, each bottleneck in RepHMS contains a $ 5 \times 5$ DWConv by default. When employing the Large Kernel (LK) strategy, the model utilizes large convolutional kernels up to $9 \times 9$ in the backbone and neck according to the GHFKS strategy. When the additional RepHConv strategy is applied, each large convolutional kernel is combined with multiple smaller kernels in parallel using re-parameterization techniques. According to the first and third rows of the Tab.~\ref{ABRepHELAN}, using large convolutional kernels results in a 0.7\% performance improvement, with the most significant gains observed in large objects, reaching 0.8\%, while the improvement for small objects is smaller. when replacing the large DWConv with RepHConv, the model’s parameter and computational cost remain unchanged, yet overall performance improves by 0.4\%, with a noticeable enhancement in small object detection. This can be summarized as follows: using large convolutional kernels effectively increases the receptive field, leading to performance gains. When combined with the RepHConv strategy, the model can optimize performance across different scales of objects. Additionally, the Cascade strategy is also a good method for achieving lossless performance improvement.

\begin{table}
	\centering
	\caption{Ablation study on RepHMS structure.}\label{ABRepHELAN}
	\renewcommand\arraystretch{1.0}	
	\setlength{\tabcolsep}{1mm}
		\begin{tabular}{ccc|cc|ccccc}
		\hline
		LK& RepHConv& Cascade &Param &FLOPs & $AP$ & $AP_{50}$& $AP_{s}$&  $AP_{m}$&  $AP_{l}$\\
    \hline
	        && & 2.0M & 7.1G & 40.5& 56.7& 21.2 &45.0 & 58.5\\
	     & &\checkmark& 2.0M & 7.1G & 40.9& 57.2& 21.5& 45.3 & 59.0\\
            \checkmark & & & 2.2M &  7.2G & 41.2&57.5& 21.4& 45.5&59.4 \\
           \checkmark  & \checkmark& & 2.2M &  7.2G & 41.6&57.8& 22.0& 45.9&59.2 \\
	  \checkmark  &   &\checkmark &2.2M&7.2G & 41.9& 58.1& 21.9& 46.0 & 59.6   \\
     \rowcolor{gray! 8} \checkmark&\checkmark&\checkmark&2.2M&7.2G & 42.3& 58.5&22.5&46.1&59.7\\ 
		\hline
		\end{tabular}
\end{table}

\subsection{Analysis of MAFPN}
In this subsection, we conducted ablation experiments on each module of MAFPN and demonstrated the plug-and-play capability of MAFPN by replacing the neck structure with different algorithms in various experiments.
\subsubsection{Ablation study on MAFPN}
The results of this experiment, as shown in Tab.~\ref{ABMAFPN}, and the default neck of the model is set to PAFPN,  which includes four C2f Blocks. Firstly, we incorporated SAF modules into the shallow layers of the backbone and neck, which resulted in a 0.5\% performance boost with an increase of 0.1M parameters and it's worth noting that through SAF, we achieved a 0.6\% improvement in performance for small targets. Secondly, with the sole addition of the AAF module, we observed an enhancement in performance specifically for objects across all scales. Next, we replaced C2f with RepHMS in the MAFPN. The model's parameter count and computational cost remained largely unchanged, while the overall performance improved by 0.5\%, demonstrating the strong effectiveness of RepHMS in the neck as well. 
Ultimately, after integrating all three strategies into MAFPN, the model’s overall performance improved by 2.1\%, with a 2.4\% increase in small object detection, demonstrating that MAFPN effectively addresses the poor performance of traditional PAFPN on small objects. Additionally, thanks to the diversified receptive fields, significant improvements were also observed in medium and large object detection.

\begin{table}
	\centering
	\caption{Ablation study on MAFPN structure.}\label{ABMAFPN}
	\renewcommand\arraystretch{0.95}	
	\setlength{\tabcolsep}{1.3mm}
		\begin{tabular}{ccc|cc|cc|ccc}
		\hline
		SAF & AAF  & RepHMS & Param &FLOPs & $AP$ & $AP_{50}$& $AP_{s}$& $AP_{m}$& $AP_{l}$\\
    \hline
	      &  & &  1.9M & 5.9G &40.2& 55.9&19.9&44.5&57.7\\
           \checkmark &  & & 2.0M & 6.2G &40.7& 56.5&20.5&44.7&58.0\\
            & \checkmark & & 2.0M & 6.8G &40.9& 56.9&20.6&44.9&58.3\\
       	&  &\checkmark & 1.9M & 5.9G &40.7& 56.8&20.3&44.6&58.5\\
        \checkmark&  &\checkmark &2.0M & 6.2G &41.1& 57.1&21.3&45.1&58.8\\
        & \checkmark  &\checkmark &2.0M & 6.8G &41.9& 58.0&21.9&45.7&59.4\\
            \checkmark &  \checkmark & & 2.2M& 7.2G&41.7&57.8&21.6&45.4&58.9\\
          
	    \rowcolor{gray! 8}\checkmark&  \checkmark & \checkmark  &2.2M&7.2G & 42.3& 58.5&22.5&46.1&59.7    \\
	   
		\hline
		\end{tabular}
\end{table}

\subsubsection{Exploring the flexibility of MAFPN}
MAFPN can be used as a plug-and-play module for other models and the results are shown in Tab.~\ref{MAFPNother}. First, we experimented with different FPN structures in MHAF-YOLO and ultimately found that only MAFPN could achieve a better balance between model parameters and performance. Then, we demonstrate the generality of this structure by replacing PAFPN with MAFPN in the mainstream single-stage detector YOLOv8n and changing the number of channels to keep the model smaller. YOLOv8n-MAFPN uses fewer epochs 
(-200 epochs) and fewer parameters and obtains a 1.7\% AP improvement, reflecting the excellent performance of MAFPN. What's more, we also verified the effectiveness of MAFPN using the two-stage detector Faster-RCNN~\cite{fasterrcnn}. By replacing the FPN in Faster R-CNN with MAFPN, we achieved a 1.2\% increase in AP with only a minimal increase in the number of parameters. In contrast, substituting FPN with PAFPN did not yield significant gains, highlighting that MAFPN maintains strong performance even in classic two-stage detectors.

\begin{table}[htp]
	\centering
	\caption{Comparative experiments of different necks on different detectors.}\label{MAFPNother}
	\renewcommand\arraystretch{0.95}	
	\setlength{\tabcolsep}{1.5mm}
		\begin{tabular}{cccccc}
		\hline
		Model &  Neck &  Param  & $AP$ & $AP_{50}$ & epoch\\
    \hline
	     \multirow{5}{*}{MHAF-YOLO} & PAFPN~\cite{pafpn} &1.9M &40.0&55.7 & 500\\
     & BIFPN~\cite{bifpn} &2.4M&41.7& 57.7&500\\
     & RepGFPN~\cite{damo} &2.6M&40.8&57.1&500\\
     & GD~\cite{goldyolo} &3.8M&42.4& 58.4&500\\
	    &  MAFPN &2.2M& 42.3&58.5&500\\
     \hline
    % \multirow{3}{*}{MHAF-YOLOn} & PAFPN & 3.1M&41.3&57.3 & 300\\
    %     & BIC &3.5M&41.7&57.9&300\\
	   %  & MAFPN &3.4M&43.1&59.4&300\\
    % \hline
     \multirow{2}{*}{YOLOv8n~\cite{yolov8}} & PAFPN & 3.2M&37.3&52.6 & 500\\
	    & MAFPN &2.6M&39.0&54.4&300\\
    \hline
	    \multirow{3}{*}{Faster-RCNN~\cite{fasterrcnn}} & FPN~\cite{fpn} &41.8M &35.7&55.2& 24\\
     & PAFPN &45.3M&35.7&55.0 &24\\
	    & MAFPN &45.0M&36.9&55.4 &24\\
		\hline
		\end{tabular}
\end{table}

% \subsection{Analysis of GAHKS}

% \begin{table}
% 	\centering
% 	\caption{Comparative Experiments on the Relationship of Convolution Kernel Sizes of RepHMS Blocks in Different Feature Layers of Backbone and Neck.}\label{tab5}
% 	\renewcommand\arraystretch{1.0}	
% 	\setlength{\tabcolsep}{0.8mm}
% 		\begin{tabular}{c|c|c|c|c|ccccc}
% 		\hline
% 		Backbone & Neck  &Method& Param &FLOPs & $AP$ & $AP_{50}$ & $AP_{s}$ & $AP_{m}$ & $AP_{l}$\\
%     \hline
%               3, 3, 3, 3   & 3, 3, 3 - 3, 3, 3 & - & 2.0M & 6.8G &41.2& 57.2&21.3&45.6&57.5 \\
%                             \hline
% 	   3, 3, 3, 3   & 9, 7, 5 - 5, 7, 9 & HKS & 2.1M & 7.0G &42.0& 58.3& 22.5& 46.2 & 58.9\\
%           9, 9, 9, 9   & 9, 7, 5 - 5, 7, 9 & HKS & 2.2M & 8.0G &42.5&58.6&23.2&47.0&58.7 \\
%               \hline
%           3, 5, 7, 9   & 3, 3, 3 - 3, 3, 3 & HKS & 2.1M & 7.0G &41.7& 57.6&21.7&46.0&58.4\\
%           3, 5, 7, 9   & 9, 9, 9 - 9, 9, 9 & HKS & 2.3M & 8.4G &42.1& 58.6& 22.7 & 46.8 & 59.0\\
%           		\hline
%           		\hline

%           9-7-5-3   & 5-7-9, 9-7-5 & GHKS & 2.1M & 8.0G &42.1& 58.1 \\
%           5-7-9-11   & 11-9-7, 7-9-11 & GAHKS & 2.3M & 7.8G &&\\
%           3-5-7-9   & 9-7-5, 5-7-9 & GAHKS & 2.2M & 7.2G && \\	   
% 		\hline
% 		\end{tabular}
% \end{table}

\subsection{Ablation study on MHAF-YOLO} 
As shown in Tab.~\ref{allMAFYOLO}, we made a series of modifications to the baseline model YOLOv10n. First, we incorporated MAFPN as the neck structure. The enhanced feature fusion led to a performance improvement of 1.6\%. However, due to the addition of extra modules and connections, the number of parameters and computational cost increased by 0.3M and 1.0G, respectively. Next, we introduced the RepHMS module. Thanks to its efficient depthwise convolutions, our model achieved a high parameter utilization rate, resulting in a performance boost of 1.1\%, while the parameter count actually decreased by 0.5M. After adding GHFKS, the performance improved by 0.7\%, with only a slight increase in the number of parameters. At this stage, the network only contained large kernel convolutions. Finally, by integrating RepHConv, the model size remained unchanged due to reparameterization, but it enriched the multi-scale representation of the MHAF-YOLO model. Ultimately, this led to a performance of 42.3\%.
\begin{table}[h]
	\centering
	\caption{Step-by-step improvements from YOLOv10n baseline to MHAF-YOLOn. The proposed setting is marked in gray.}\label{allMAFYOLO}
	\renewcommand\arraystretch{1.0}	
	\setlength{\tabcolsep}{3mm}
		\begin{tabular}{c|cc|c}
		\hline
		Model  &Param &FLOPs & $AP$  \\
    \hline 
            YOLOv10n baseline&   2.3M & 6.7G & 38.5 \\
	    + MAFPN&   2.6M ($+$0.3) & 7.7G ($+$1.0) & 40.1 ($+$1.6) \\
           + RepHMS&   2.1M ($-$0.5) & 7.0G ($-$0.7) &41.2 ($+$1.1)\\
           + GHFKS&   2.2M ($+$0.1) & 7.2G ($+$0.2) & 41.9 ($+$0.7) \\
           \rowcolor{gray! 15} + RepHConv&   2.2M  & 7.2G  & 42.3 ($+$0.4) \\

		\hline
		\end{tabular}
\end{table}
% \begin{table}[h]
% 	\centering
% 	\caption{Ablation study on all modules of MHAF-YOLO.}\label{allMAFYOLO}
% 	\renewcommand\arraystretch{1.0}	
% 	\setlength{\tabcolsep}{3mm}
% 		\begin{tabular}{cccccc}
% 		\hline
% 		MAFPN & RepHMS& GAHKS &Param &FLOPs & $AP$ \\
%     \hline
% 	      &  & & 4.3M & 10.9G & 37.7 \\
%             \checkmark&  && 4.8M  &  12.1G & 39.8   \\
% 	    \checkmark& \checkmark && 3.6M & 10.4G & 40.9\\
% 	    \checkmark&  \checkmark &\checkmark &3.8M &10.5G & 42.4    \\
	   
% 		\hline
% 		\end{tabular}
% \end{table}

\begin{table*}
	\small
	\centering
	\caption{Comparison with state-of-the-art real-time object detectors. $^\dag$ means the results of models with the original one-to-many training using NMS. $^\ddag$ represents that self-distillation method is utilized, and * refers to train with pertained backbone.}\label{SOTA}
	\renewcommand\arraystretch{1.02}
    \resizebox{\textwidth}{!}{
		\begin{tabular}{cccccccc}
		\hline
		Methods & $AP^{val}$& $AP^{val}_{50}$&  $AP^{s}$&  $AP^{m}$&  $AP^{l}$& Params& FLOPs\\
		\hline
		YOLOv6-n$^\ddag$ ~\cite{yolov6} &   37.5  & 53.1 & 17.8&{\itshape} 41.8&55.1& 4.7M&11.4G\\
		YOLOv7-t~\cite{yolov7}  &  37.4  & 55.2 & 19.0& 41.8& 52.6& 6.2M & 13.7G\\
		% DAMO-YOLO-t & 42.0 & 58.0 & 23.0 & 46.1 & 58.5 & 8.5M & 18.1G\\
        RTMDet-t*~\cite{rtmdet}  &   41.0 & 57.4 & 20.7& 45.3& 58.0 & 4.9M & 16.2G \\
        YOLOv8-n~\cite{yolov8}  &   37.3  & 52.6 & 18.5& 41.0& 53.5& 3.2M & 8.7G\\
        Gold-YOLO-n~\cite{goldyolo}  &   39.9  & 55.9 & 19.1& 44.3& 57.8& 5.6M & 12.1G\\
        YOLOv9-t~\cite{yolov9} & 38.3 & 53.1 & 18.6 & 42.3 & 54.7 & 2.0M & 7.7G\\
        MAF-YOLO-n~\cite{mafyolo}  &   42.4  & 58.9 & 22.0 & 46.5 & 59.4 & 3.8M& 10.5G\\
        YOLOv10-n~\cite{yolov10} & 38.5 / 39.5$^\dag$ & 53.8 / 55.0$^\dag$& 18.9 & 42.4 & 54.6 & 2.3M & 6.7G\\
        YOLO11-n~\cite{mafyolo}  &   39.5  & 55.2 & 19.9 & 43.3 & 56.8& 2.6M& 6.5G\\
        YOLOv12-n~\cite{yolov12}  &   40.6  & 56.7 & 20.2 & 45.2 & 58.4& 2.6M& 6.5G\\
		\rowcolor{gray! 15} MHAF-YOLO-lite-n  &   38.5 / 39.5$^\dag$ & 53.7 / 55.0$^\dag$& 18.5 & 42.9 & 53.8 & 1.4M& 4.7G\\
		\rowcolor{gray! 15} MHAF-YOLO-n  &   42.3 / 43.2$^\dag$  & 58.5 / 59.4$^\dag$ & 22.5 & 46.1 & 59.7 & 2.2M& 7.2G\\
        \rowcolor{gray! 15} MHAF-YOLO-n*  &   43.1  & 59.3 & 22.3 & 47.5 & 60.5 & 2.2M& 7.2G\\
		\hline
    \
		% YOLOXs~\cite{yolox} & 40.7&59.6&23.9&45.2&53.8& 9.0M& 26.8G\\
		YOLOv6-s$^\ddag$ & 45.0 & 61.8 & 24.3& 50.2& 62.7& 18.5M& 45.3G\\
  	YOLOv7-s-AF & 45.1 &61.8 & 25.7& 50.2& 61.2& 11.0M& 28.1G\\
        DAMO-YOLO-s~\cite{damo} & 46.0 & 61.9 & 25.9 & 50.6 & 62.5 & 12.3M & 37.8G\\
        RTMDET-s* & 44.6 &61.7 &24.2&49.2&61.9&8.9M&29.6G\\
		YOLOv8-s &  44.9 & 61.8 & 25.7& 49.9& 61.0&11.2M&28.6G\\
        YOLOMS-s~\cite{yoloms} &  46.2  & 63.7 & 26.9& 50.5& 63.0&8.1M&31.2G\\
        Gold-YOLO-s & 45.4 & 62.5 & 25.3 & 50.2 & 62.6 & 21.5M & 46.0G\\
        % RT-DETR-s & 46.5 & 63.8 & - & - & - & 20.0M & 60.0G\\
        YOLOv9-s &  46.8  & 63.4 & 26.6& 56.0& 64.5&7.2M&26.7G\\
		MAF-YOLO-s & 47.4 & 64.3 & 27.8& 51.9& 64.8& 8.6M& 25.5G\\
        RT-DETRv2-s & 48.1 & 65.1 & - & - & - & 20.0M & 60.0G\\
        YOLOv10-s & 46.3 / 46.8$^\dag$& 63.0 / 63.7$^\dag$& 26.8 & 51.0 & 63.8 & 7.2M & 21.6G\\
        YOLO11-s~\cite{mafyolo}  &   47.0  & 63.9 & 29.0 & 51.7 & 64.5 & 9.4M& 21.5G\\
        YOLOv12-s  &   48.0  & 65.0 & 29.8 & 53.2 & 65.5& 9.3M& 21.4G\\
        \rowcolor{gray! 15} MHAF-YOLO-s &   48.9 / 49.1$^\dag$ & 65.9 / 66.1$^\dag$& 30.2 & 53.9 & 65.1 & 7.1M& 25.3G\\
        \rowcolor{gray! 15} MHAF-YOLO-s*  &   49.4  & 66.5 & 30.6 & 54.1 & 65.9 & 7.1M& 25.3G\\
	    \hline
		YOLOv6-m* &  50.0  &66.9 & 30.6& 55.4&67.3& 34.9M& 85.8G\\
		YOLOv7 & 51.2 & 69.7 & 31.8 & 55.5 & 65.0 & 36.9M & 104.7G\\
        DAMO-YOLO-m$^\ddag$ &  50.4 & 67.2 &25.9 &50.6 &62.5 &28.2M&61.8G\\
        RTMDet-m & 49.3 & 66.9 & 30.5 & 53.6 & 66.1 & 24.7M & 78.6G\\
        YOLOv8-m &  50.2 & 67.2 & 32.1& 55.7& 66.5&25.9M&78.9G\\
        YOLOMS &  51.0 & 68.6 & 33.1& 56.1& 66.5&22.2M&80.2G\\
        Gold-YOLO-m & 49.8 & 67.0 & 32.3 & 55.3 & 66.3 & 41.3M & 87.5G\\
        YOLOv9-m & 51.4 & 68.1 & 33.6 & 57.0 & 68.0 & 20.0M & 76.3G\\
	MAF-YOLO-m & 51.2  & 68.5 & 33.2 & 56.3 & 67.5 & 23.7M & 76.7G\\
        RT-DETRv2-m~\cite{rtdetrv2} & 51.9 & 69.9 & - & - & - & 36.0M & 100.0G\\
        YOLOv10-m & 51.1 / 51.3$^\dag$& 68.1 / 68.5$^\dag$& 33.8 & 56.5 & 67.0 & 15.4M & 59.1G\\
        YOLO11-m~\cite{mafyolo}  &   51.5  & 68.5 & 33.4 & 57.1 & 67.9& 20.1M& 68.0G\\
        YOLOv12-m  &   52.5  & 69.6 & 35.7 & 58.3 & 68.8& 20.2M& 67.5G\\
        \rowcolor{gray! 15} MHAF-YOLO-m & 52.7 / 52.8$^\dag$  & 69.5 / 69.7$^\dag$ & 35.2 & 57.6 & 67.6 & 15.3M&65.2G\\
    \hline
    %TOOD-R101& 46.7  &64.6 & 28.9& 49.6& 57.0& 51.2M& 249G&1333&24\\
    % ConvNeXt-T(Mask R-CNN)~\cite{convnext}& 46.2 / - &68.1 / - & 30.1& 49.5& 59.5& 48.1M& 262G\\
    Faster-RCNN-R101~\cite{fasterrcnn}& 39.8 &60.1 & 22.5& 43.6& 52.8& 60.7M& 255G\\
    Swin-T(Cascade-Mask-RCNN)~\cite{swin, cascade}& 50.5 &69.3 & -& -& -& 86.0M& 745G\\
    TOOD-R101~\cite{tal}& 46.1 &63.8 & 30.2& 49.7& 58.8& 51.1M& 249G\\
    ConvNeXt-T~\cite{convnext}& 46.2 &68.1 & 30.1& 49.5& 59.5& 48.1M& 262G\\
    Deformable DETR~\cite{deformable} &  46.2 &65.2 & 28.8& 49.2& 61.7& 40.1M& 173G\\
    ViTDet-ViT-B~\cite{vit} &  51.6 &72.1 & 35.3& 55.6& 66.4& 111.0M& 840G\\
    DINO-4scale-R50~\cite{dino} &  50.4 &68.3 & 33.3& 53.7& 64.8& 47.7M& 279G\\
    
		% YOLOv5m & {\itshape} 45.4$\%$ &{\itshape} 64.1$\%$&{\itshape} 21.2M&{\itshape} 49.0G\\
		% YOLOv6m & {\itshape} 50.0  &{\itshape} 66.9 &{\itshape} 30.6&{\itshape} 55.4&{\itshape} \textbf{67.3}&{\itshape} 34.9M&{\itshape} 85.8G\\
		% YOLOv8m & {\itshape} 50.2  &{\itshape} 67.2 &{\itshape} 32.1&{\itshape} 55.7&{\itshape} 66.5&{\itshape} \textbf{25.9M}&{\itshape} 78.9G\\
		% \textbf{MA-YOLOm}  &  {\itshape} \textbf{50.8}  &{\itshape} \textbf{68.2} &{\itshape} \textbf{33.0}&{\itshape} \textbf{56.5}&{\itshape} 67.2&{\itshape} 28.4M&{\itshape} \textbf{72.7G}\\
    \hline

	\end{tabular}}
\end{table*}

% \subsection{Comparison with State-of-the-Arts}
\subsection{Detector on Different Datasets with Different tasks}
% To demonstrate the effectiveness and generalization of our proposed model and modules, we conducted extensive experiments with MHAF-YOLO across various datasets and tasks, including object detection on the COCO and VOC datasets, instance segmentation on the COCO-seg dataset, and rotated object detection on the DOTA-v1.0 dataset. We also performed comprehensive comparisons with state-of-the-art (SOTA) methods or the latest approaches in these domains.
\subsubsection{Comparison of Real-Time Detectors on COCO}
\begin{figure}[H]
	
	\begin{minipage}[h]{1.0\linewidth}
		\centering
		\centerline{\includegraphics[width=12cm]{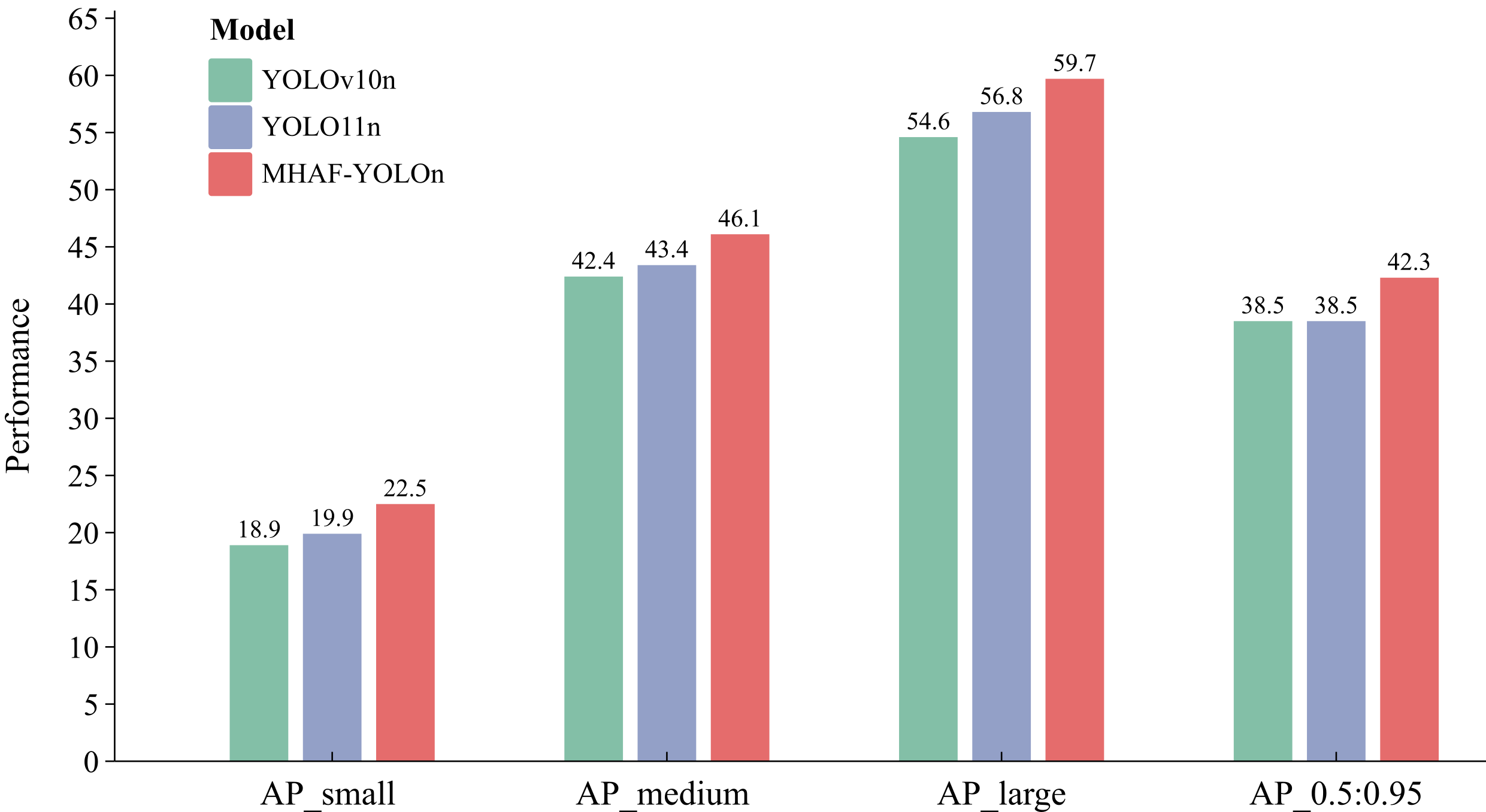}}
		%  \vspace{2.0cm}
		% \centerline{(a) Result 1}\medskip
	\end{minipage}
	\caption{Multi-scale performance comparison chart.}
	\label{example}
\end{figure}

\begin{figure}
\centering
\includegraphics[width=1\linewidth]{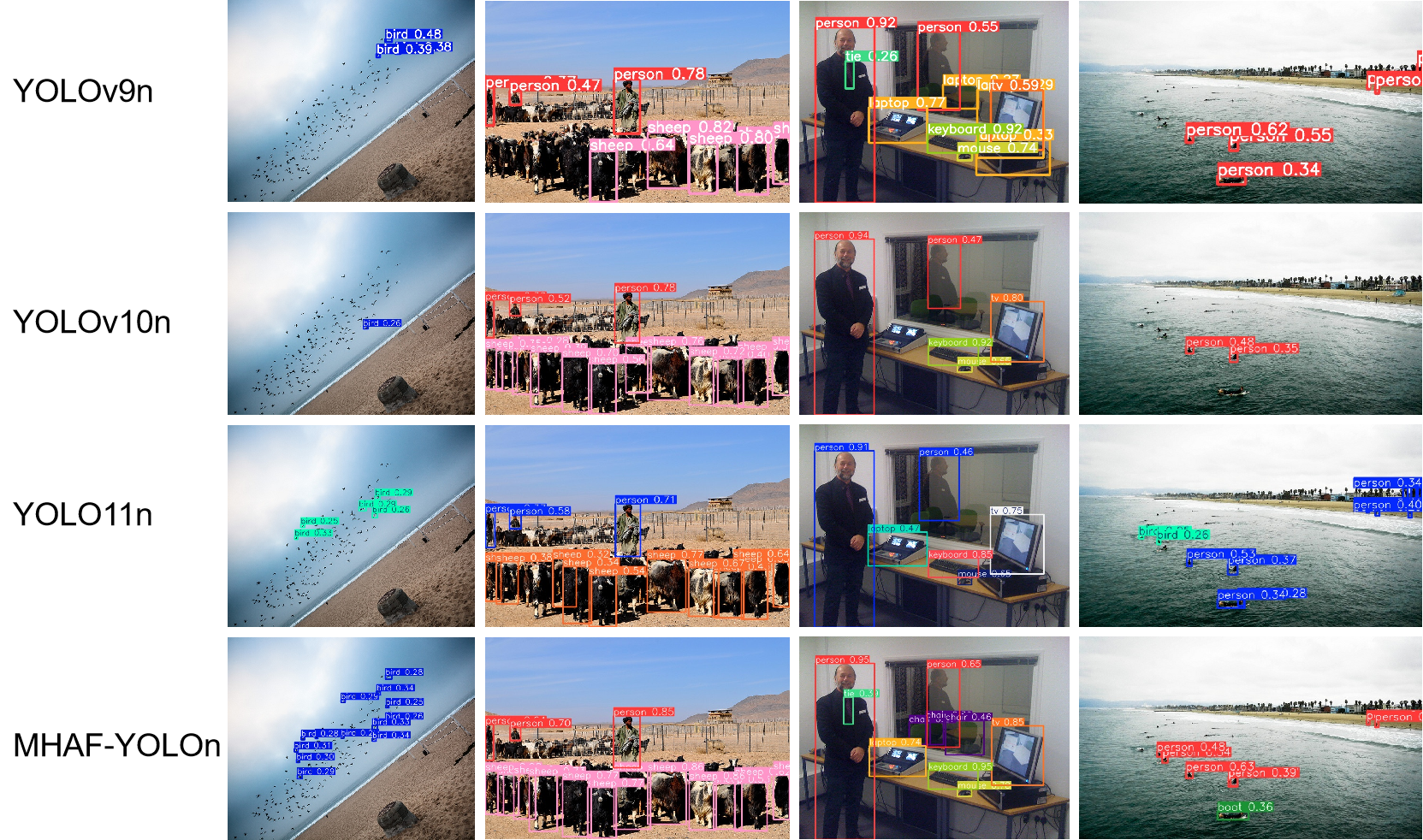}
\caption{The detection results of YOLOv9n, YOLOv10n, YOLO11n, MHAF-YOLOn.}
\label{result_fig}
\end{figure}

Tab.~\ref{SOTA} shows the comparison results between our proposed MHAF-YOLO and other state-of-the-art real-time object detectors. We first compare MHAF-YOLO with our baseline models, i.e., YOLOv10. On N / S / M  three variants, our model achieves 3.8\% / 2.6\% / 1.6\% AP
improvements, with fewer parameters. Compared with other YOLOs, MHAF-YOLO also exhibits superior trade-offs between accuracy and computational cost. For example, compared with Gold-YOLO, MHAF-YOLO has shown an extraordinary efficiency in utilizing parameters, The number of parameters in MHAF-YOLO N / S / M is 60\% / 67\% / 63\% less than that of Gold-YOLO, but it still achieves a performance gain of 2.4\% / 3.5\% / 2.9\%. Our model also has a significant advantage for even smaller-scale models, Compared to the YOLOv6-n, YOLOv7-t, YOLOv8-n, YOLOv9-t, YOLOv10-n, The MHAF-YOLO-lite-nano model demonstrates remarkable lightweight potential by achieving a significant reduction of 30\% to 77\% in parameters and 30\% to 66\% in computational requirements, while maintaining comparable Average Precision (AP) scores. Compared to the latest YOLO11, our three-scale models have fewer parameters yet outperform YOLO11-n, YOLO11-s, and YOLO11-m by 2.8\%, 1.9\%, and 1.2\%, respectively. This underscores the excellence of our model in achieving high performance with minimal resource consumption, making it ideal for applications where efficiency and compactness are paramount. In addition, we present several two-stage and transformer-based detectors where our model demonstrates superior performance and is more lightweight. In addition, thanks to the MAFPN and multi-scale receptive field modules, our model significantly outperforms others in detecting multi-scale objects. As shown in Fig.~\ref{example}, we present the detection performance for targets of three different scales in a bar chart, where MHAF-YOLO consistently outperforms both YOLOv10 and YOLO11 across all metrics. And some detection results of different algorithms on the COCO validation set are shown in Fig.~\ref{result_fig}.

\subsubsection{Comparison of Real-Time Detectors on VOC} We train the mainstream YOLO series models on the VOC dataset for 300 epochs and report system-level comparisons in Tab.~\ref{tab8}, which shows that our model is 2\% and 1.7\% AP higher than the baseline YOLOv10n and YOLOv10s, respectively. Compared to the best-performing YOLOv9, MHAF-YOLO still shows a significant advantage in the $AP_{50}$ metric. MHAF-YOLO n/s also achieved performance gains of 1.4\% and 0.9\% over YOLO11 n/s, demonstrating the superior generalization of our model on the VOC dataset.
\begin{table}[H]
	\centering
         \small
	\caption{Comparison with state-of-the-art YOLO real-time object detectors on VOC.}\label{tab8}
	\renewcommand\arraystretch{1}	
	\setlength{\tabcolsep}{5mm}
		\begin{tabular}{ccccc}
		\hline
		Methods & $AP^{val}$& $AP^{val}_{50}$ &Param &FLOPs \\
    \hline
        YOLOv7-t~\cite{yolov7}  &  50.8  & 76.3 & 6.1M & 13.2G\\
        YOLOv8-n~\cite{yolov8}  & 58.9 & 79.8 & 3.0M & 8.1G\\
        Gold-YOLO-n~\cite{goldyolo}  & 58.4 & 79.8 & 5.6M & 12.1G\\
        YOLOv9-t~\cite{yolov9} & 61.9 & 80.6 & 2.6M & 10.7G\\
        YOLOv10-n~\cite{yolov10} & 59.9 & 79.9 & 2.2M & 6.5G\\
        YOLO11-n~\cite{yolo11} & 60.5 & 80.3 & 2.6M & 6.5G\\
		% \rowcolor{gray! 8} YOLO  &   38.5  & 53.7 & 18.5 & 42.9 & 53.8 & 1.4M& 4.7G\\
		\rowcolor{gray! 8} MHAF-YOLO-n  & 61.9 & 81.3 & 2.2M & 7.1G\\
		\hline
	YOLOv8-s & 64.0 & 83.6 & 11.1M & 28.5G\\
        Gold-YOLO-s & 62.6 & 82.8 & 21.5M & 46.1G\\
        YOLOv9-s &  65.4  & 83.4 & 9.6M & 38.8G\\
        YOLOv10-s & 63.9 & 83.3 & 7.2M & 21.6G\\
        YOLO11-s & 64.7 & 83.8 & 9.4M & 21.5G\\
        \rowcolor{gray! 8} MHAF-YOLO-s & 65.6 & 84.7 & 7.1M & 25.1G\\
	   
		\hline
        Faster-RCNN-R101~\cite{fasterrcnn} & 48.6 & 77.2 & 60.4M & 251G\\
        Cascade-RCNN-R101~\cite{cascade} & 51.1 & 75.7 & 88.2M & 278G\\
        Dino-4scale-R50~\cite{dino} & 65.3 & 84.6 & 47.6M & 244G\\
        \hline
		\end{tabular}
\end{table}

\subsubsection{Results on Semantic Segmentation}
In addition to the object detection task, we have also conducted a performance evaluation of the instance segmentation task on the COCO dataset. 
% To retain more characteristics of MHAF-YOLO, we kept most of the structure and training hyperparameters unchanged, merely replacing the output detection head with a mask feature head similar to that of YOLO11 and substituting the corresponding segmentation loss. Furthermore, we have adopted a one-to-many label assignment strategy and retained the NMS operation in the post-processing stage.
As shown in Tab.~\ref{tab10}, our nano/small model outperforms YOLO11 n/s in segmentation AP by 3.6\% and 2.2\%, respectively, while having fewer parameters and lower computational costs. This demonstrates that MHAF-YOLO also shows promising potential in instance segmentation tasks.
\begin{table}[H]
	\centering
         \small
	\caption{ Comparison of MHAF-YOLO-seg with other instance segmentation methods.}\label{tab10}
	\renewcommand\arraystretch{1}	
	\setlength{\tabcolsep}{1.2mm}
		\begin{tabular}{ccccccc}
		\hline
		Methods &Input shape& $AP^{bbox}$& $AP^{seg}$& Params& FLOPs& epochs\\
  \hline
        HTC-R101~\cite{htc}  & 800-1333 &44.8  & 39.6 & 99.3M &1803G & 20\\
        SOLOv2-R50~\cite{solov2}  & 800-1333 &-  & 37.5 & 46.4M &253.5G & 36\\
        Cascade-RCNN-R50~\cite{cascade}  & 800-1333 &44.3  & 38.5 & 77.1M &403.6G & 36\\
        RTMDet-ins-R50-FPN~\cite{rtmdet}  & 800-1333 & 45.3 & 39.7 & 35.9M & 295.2G & 36\\
        
  	\hline
        YOLOv5-n-seg~\cite{yolov5}  & $640^2$ &27.6  & 23.4 & 2.0M &7.1G & 300\\
        YOLOv5-s-seg  & $640^2$ & 37.6 & 31.7 & 7.6M & 26.4G & 300\\
        YOLOv6-n-seg~\cite{yolov6}  & $640^2$ & 35.3 & 31.2 & 4.9M & 7.0G & 300\\
        RTMDet-tiny-ins  & $640^2$ & 40.5 & 35.4 & 5.6M & 23.6G & 300\\
        YOLOv8-n-seg~\cite{yolov8}  & $640^2$ & 36.7 & 30.5 & 3.4M & 12.6G & 500\\
        YOLO11-n-seg~\cite{yolo11}  & $640^2$ & 38.9 & 32.0 & 2.9M & 10.4G & 500\\
        \rowcolor{gray! 8} MHAF-YOLO-n-seg & $640^2$ & 42.5 & 35.0  & 2.4M & 14.8G & 500\\
        \hline
        YOLOv5-m-seg  & $640^2$ & 45.0 & 37.1 & 22.0M & 70.8G & 300\\
        YOLOv6-s-seg  & $640^2$ & 44.0 & 38.0 & 19.6M & 27.7G & 300\\
        RTMDet-s-ins & $640^2$ & 44.0 & 38.7 & 10.2M & 43.0G & 300\\
        YOLOv8-s-seg & $640^2$ & 44.6 & 36.8 & 11.8M & 42.6G & 500\\
        YOLO11-s-seg  & $640^2$ & 46.6 & 37.8 & 10.1M & 35.5G & 500\\
        \rowcolor{gray! 8} MHAF-YOLO-s-seg & $640^2$ & 48.8 & 39.7 & 7.8M & 40.4G & 500\\
	\hline
		\end{tabular}
\end{table}

\subsubsection{Results on Rotated Object Detection}
\begin{table}
\centering
\normalsize
\small
\caption{A comparison of MHAF-YOLO-obb with previous rotated object detection methods is conducted on the DOTA-v1.0 test set, focusing on the number of parameters, FLOPs, and accuracy. IN, COCO, and MAE respectively represent ImageNet pretraining, COCO pretraining, and MAE unsupervised pretraining~\cite{maedet}. MS indicates multi-scale training and testing.}\label{rotate}
	\renewcommand\arraystretch{1}	
	\setlength{\tabcolsep}{1.7mm}
    
\begin{tabular}{c|c|c|c|c|c|c}
		 \hline
        Method & Pretrain & Backbone & MS & Params & FLOPs & mAP \\ 
        \hline
        \textit{Two-stage} &&&&&&\\
        \hline
        Oriented RCNN~\cite{oriented}&  IN  & PKINet-T~\cite{pkinet} & \ding{55} & 4.1M & 45.4G & 77.87 \\
        Oriented RCNN&  IN  & PKINet-S & \ding{55} & 13.7M & 70.2G & 78.39 \\
        Oriented RCNN&  IN  & R50 & \checkmark & 41.1M & 398G & 80.87 \\
        KFIoU~\cite{kfiou}&  IN  & Swin-T~\cite{swin} & \checkmark & 58.8M & 416G & 80.93 \\
        RVSA~\cite{rvsa}&  MAE  & ViTAE-B~\cite{vitae} & \checkmark & 114.4M & 828G & 81.24 \\

        Oriented RCNN&  IN  & LSKNet-T~\cite{lsknet} & \checkmark & 21.0M & 248G & 81.37 \\
		 \hline
                \textit{One-stage} &&&&&&\\
        \hline
        PPYOLOE-R-s~\cite{ppyoloe} & IN  & CRN-s &  \ding{55}  & 8.1M & 43.5G & 73.82 \\
        PPYOLOE-R-s & IN  & CRN-s &  \checkmark  & 8.1M & 43.5G & 79.42 \\
        YOLOv8-s-obb & COCO  & YOLOv8-s & \checkmark & 11.4M & 76.3G & 79.50 \\
        YOLOv8-m-obb & COCO  & YOLOv8-m & \checkmark & 26.4M & 208.6G & 80.50 \\
        RTMDet-R-t~\cite{rtmdet} & IN  & RTMDet-t &  \ding{55}  & 4.9M & 40.9G & 75.36 \\
        RTMDet-R-t & IN  & RTMDet-t & \checkmark & 4.9M & 40.9G & 79.82 \\
        RTMDet-R-s & IN  & RTMDet-t & \ding{55} & 8.9M & 75.2G & 76.93 \\
        RTMDet-R-s & IN  & RTMDet-s & \checkmark & 8.9M & 75.2G & 79.98 \\
        RTMDet-R-m & IN  & RTMDet-m & \ding{55} & 24.7M & 199.5G & 78.24 \\
        RTMDet-R-m & IN  & RTMDet-m & \checkmark & 24.7M & 199.5G & 80.26 \\
        RTMDet-R-l & IN  & RTMDet-l & \ding{55} & 52.3M & 410G & 78.85 \\
        RTMDet-R-l & IN  & RTMDet-l & \checkmark & 52.3M & 410G & 80.54 \\
        YOLO11-n-obb~\cite{yolo11} & COCO  & YOLO11-n & \checkmark & 2.7M & 17.2G & 78.40 \\
        YOLO11-s-obb & COCO  & YOLO11-s & \checkmark & 9.7M & 57.9G & 79.50 \\
        YOLO11-m-obb & COCO  & YOLO11-m & \checkmark & 20.9M & 183.5G & 80.90 \\
        % YOLO11-x-obb & COCO  & YOLO11-m & \checkmark & 58.8M & 520.2G & 81.30 \\
          	\hline
        \rowcolor{gray! 8}MHAF-YOLO-n-obb & \ding{55} & Ours-nano & \ding{55} & 2.2M & 19.3G & 77.95 \\
        \rowcolor{gray! 8}MHAF-YOLO-n-obb & IN & Ours-nano & \ding{55} & 2.2M & 19.3G & 79.36 \\
        \rowcolor{gray! 8}MHAF-YOLO-n-obb & \ding{55} & Ours-nano & \checkmark & 2.2M & 19.3G & 79.77 \\
        % \rowcolor{gray! 8}MHAF-YOLO-n-obb & COCO & Ours-nano & \checkmark & 2.2M & 19.3G & 80.21 \\
        \rowcolor{gray! 8}MHAF-YOLO-s-obb & \ding{55} & Ours-small & \ding{55} & 7.3M & 67.3G & 79.52 \\
        \rowcolor{gray! 8}MHAF-YOLO-s-obb & IN & Ours-small & \ding{55} & 7.3M & 67.3G &  80.31\\
        \rowcolor{gray! 8}MHAF-YOLO-s-obb  & \ding{55} &  Ours-small & \checkmark & 7.3M & 67.3G & 81.10 \\
                  	\hline
        % \rowcolor{gray! 8}MHAF-YOLO-s-obb  & COCO &  Ours-small & \checkmark & 7.3M & 67.3G &  \\
	  \end{tabular}
\end{table}
In Tab.~\ref{rotate}, we compare MHAF-YOLO-obb with previous state-of-the-art methods on the DOTA-v1.0 dataset. In single-scale training and inference, the model faces the more complex issue of small object detection. MHAF-YOLO-n-obb and MHAF-YOLO-s-obb achieve 79.36\% and 80.31\% mAP, respectively, outperforming nearly all previous single-scale methods. Specifically, our nano achieves a 0.51\% performance improvement with only about 1/20th of the parameters and computational cost compared to the RTMDet-R-l model, ultimately reaching 79.36\% AP. Without ImageNet pretraining, our small model also reaches an AP of 79.52\%, surpassing the O-RCNN~\cite{oriented} method with the state-of-the-art backbone PKINet-S~\cite{pkinet} by 1.13\%, while reducing the parameter count by 46.7\%. In multi-scale training and testing, YOLOv8 and YOLO11 employs the COCO pretraining strategy, significantly boosting performance. Under the same training strategy, our MHAF-YOLO-n-obb, without pretraining, still outperforms YOLO11-n-obb and YOLO11-s-obb by 1.37\% and 0.27\%, respectively. Our MHAF-YOLO-s-obb achieves an AP of 81.10\% in multi-scale settings, almost matching the most advanced rotated object detection methods. For example, RVSA~\cite{rvsa} requires an extremely large model and pretraining on a massive dataset to barely exceed 81 AP. When compared to the state-of-the-art method LSKNet-T~\cite{lsknet}, our model shows clear advantages in both parameter count and computational cost. We believe that our model can achieve even more advanced performance with stronger pretraining and further optimizations in rotated object detection in the future.

\section{Conclusions}
In this paper, we reviewed the commonly used PAFPN architecture in YOLO networks and critically analyzed its limitations in feature fusion. Building on this, we proposed a robust multi-scale feature fusion network, MAFPN, which is both efficient and versatile. It can be seamlessly integrated into any object detector to enhance performance. MAFPN incorporates two specialized modules in its shallow and deep layers, SAF and AAF, respectively. The SAF module strategically integrates more information from the shallow backbone, significantly improving localization accuracy and small-object detection capability. The AAF module employs a richer connectivity mechanism, enabling extensive interaction of multi-scale feature information within the neck. As the network undergoes iterative updates, these features mutually complement and refine each other, ultimately producing more informative gradients at the output layer. We also examined the importance of multi-scale receptive fields from both global and local perspectives. Globally, we introduced the GHFKS mechanism, which adaptively adjusts convolution kernel sizes based on the target layer dimensions, progressively expanding the network’s overall receptive field. Locally, we designed re-parameterized heterogeneous convolutions to mitigate the loss of small-object information caused by excessively large kernels. Based on the above innovations, we developed the MHAF-YOLO model, which achieves exceptional parameter efficiency and state-of-the-art performance. Compared to similar models, our approach demonstrates superior performance in three benchmark datasets, COCO, VOC, and DOTA-v1.0, achieving state-of-the-art results in object detection, instance segmentation, and rotated object detection tasks. We hope this work provides new insights for building higher-precision real-time object detectors.

Although MHAF-YOLO achieves high accuracy with comparable computational costs, its inference speed still lags behind cutting-edge models such as YOLOv10 and YOLO11. This is primarily due to the relative complexity of MAFPN and the inefficiency of large-kernel depthwise convolutions. In future work, we aim to maintain high accuracy while optimizing the model’s inference speed to better meet the demands of industrial applications.
\section*{Declarations}
\begin{itemize}
\item Conflict of interest/Competing interests (check journal-specific guidelines for which heading to use): Authors declare no competing interests that are directly or indirectly related to the work submitted for publication.
\item Ethics approval and consent to participate: Not applicable.
\item Consent for publication: Not applicable.
\item Data availability: The datasets used in this study are publicly available and include: COCO \cite{coco}, VOC \cite{voc} and DOTA-v1.0 \cite{dota}.
\item Materials availability: Not applicable.
\item Code availability: Source code is available at \url{ https://github.com/yang-0201/MHAF-YOLO}
% \item Author contribution: All authors contributed to the study conception and design. Material preparation, data collection, and analysis were performed by [Huihuang Zhang] and [Haigen Hu]. The methods of conceptualization and code writing were carried out by [Huihuang Zhang]. The first draft of the manuscript was written by [Huihuang Zhang] and [Xiaoqin Zhang]. Grammar and format were edited by [Bin Cao]. All authors commented on previous versions of the manuscript. All authors read and approved the final manuscript.
\end{itemize}

\section*{Declaration of competing interest}
The authors declare that they have no competing interests that are directly or indirectly related to the work submitted for publication.
\section*{CRediT authorship contribution statement}
\textbf{Zhiqiang Yang}: Conceptualization, Methodology, Investigation, Visualization, Writing – original draft. \textbf{Qiu Guan}: Conceptualization, Methodology, Supervision, Funding acquisition. \textbf{Zhongwen Yu}: Methodology, Conceptualization, Resources. \textbf{Xinli Xu}: Writing – review \& editing, Investigation. \textbf{Haixia Long}: Writing – review \& editing. \textbf{Sheng Lian}: Writing – review \& editing, Validation. \textbf{Haigen Hu}: Formal analysis.  \textbf{Ying Tang}: Visualization, Writing – review \& editing.

\section*{Data availability}
Data will be made availability on request.
\section*{Declaration of Generative AI and AI-assisted technologies in the writing process}
During the preparation of this work, the author(s) used GPT-4o to improve readability and language expression. After using this tool, the author(s) reviewed and edited the content as needed and take(s) full responsibility for the content of the publication.

\section*{Acknowledgement}
The authors would like to thank the referees for the helpful comments and suggestions. Portions of this work were presented at the 7th Chinese Conference on Pattern Recognition and Computer Vision (PRCV 2024 Oral), entitled "Multi-branch Auxiliary Fusion YOLO with Re-parameterization Heterogeneous Convolutional for Accurate Object Detection". This work is supported in part by National Natural Science Foundation of China (U20A20171, 72192823, 62373324, 62271448), Natural Science Foundation of Zhejiang Province (LY21F020027, LZ23F020010, 61972355) and Key Programs for Science and Technology Development of Zhejiang Province (2022C03113).
% \appendix
% \section{Supplementary Materials}
% This supplementary material offers detailed insights and further analysis regarding MAF-YOLO implementation. Firstly, we provide an elaborate description of parameter configurations both pre-training and during training. Following that, we present the detailed network model structure of MAF-YOLOn. Lastly, we conduct additional visual analysis of the model's effective receptive field.
% \subsection{Implementation Details}

\bibliographystyle{elsarticle-num} 
\bibliography{ref}
\small

\end{document}